\documentclass[journal]{IEEEtran}

\usepackage{times}
\usepackage{epsfig}
\usepackage{amsmath}
\usepackage{amssymb}
\usepackage{multirow}
\usepackage{booktabs}
\usepackage{graphicx}
\usepackage{float}
\usepackage[colorlinks]{hyperref}
\usepackage{cite}
\usepackage{arydshln}
\usepackage{balance}
\usepackage{hyperref}
\usepackage{cleveref}     

\usepackage[table]{xcolor}
\usepackage{xcolor}
\usepackage{makecell}
\usepackage{bbding}
\usepackage{microtype}
\usepackage{units}
\usepackage{bm}
\usepackage{enumitem}
\usepackage[normalem]{ulem}
\usepackage{svg}
\usepackage{pdfpages}
\usepackage{multirow}

\usepackage[defaultcolor=black]{changes}
\definechangesauthor[name={author}, color=red]{A}
\useunder{\uline}{\ul}{}

\definecolor{brickred}{RGB}{200, 50, 50}  
\definecolor{blueish}{RGB}{0, 128, 0} 
\usepackage{hyperref}
\hypersetup{hypertex=true,
    colorlinks=true,
    linkcolor=brickred,
    anchorcolor=blueish,
    citecolor=blueish}

\newcommand{\blue}[1]{\textcolor{blue}{#1}}
\newcommand{\figtwo}[2]{\hyperref[#1]{~\ref{fig:mov_4}~(#2)}}
\newcommand{\figfive}[2]{\hyperref[#1]{~\ref{fig:ipc}~(#2)}}
\newcommand{\figeight}[2]{\hyperref[#1]{~\ref{fig:viz}~(#2)}}

\def\VspaceFa{\vspace{-0.40cm}}
\def\VspaceFb{\vspace{-0.30cm}}

\def\VspacePa{\vspace{-0.30cm}}

\def\VspacePc{\vspace{-0.15cm}}

\def\VspaceSen{\vspace{0.10cm}}

\begin{document}

\title{ 
NanoHTNet: Nano Human Topology Network for Efficient 3D Human Pose Estimation
}
\author{
Jialun Cai, Mengyuan Liu$^{\dagger}$, Hong Liu, Shuheng Zhou, Wenhao Li
\thanks{$^{\dagger}$ Corresponding author.}
\thanks{
J. Cai was with the State Key Laboratory of General Artificial Intelligence, Peking University, Shenzhen Graduate School. He is now with Ant Group (Email: caijialun.cjl@antgroup.com).

M. Liu$^{\dagger}$ and H. Liu are with the State Key Laboratory of General Artificial Intelligence, Peking University, Shenzhen Graduate School (Email: \{liumengyuan, hongliu\}@pku.edu.cn). 

S. Zhou is with the Ant Group (Email: shuheng.zsh@antgroup.com). 

W. Li is with the School of Computer Science and Engineering, Nanyang Technological University, Singapore (E-mail: wenhao.li@ntu.edu.sg).

This work was supported by National Natural Science Foundation of China (No. 62473007), Natural Science Foundation of Guangdong Province (No. 2024A1515012089), Shenzhen Innovation in Science and Technology Foundation for The Excellent Youth Scholars (No. RCYX20231211090248064), and Ant Group Research Fund.}}

\markboth{IEEE Transactions on Image Processing}
{Cai \MakeLowercase{\textit{et al.}}: 
NanoHTNet: Nano Human Topology Network for Efficient 3D Human Pose Estimation}

\maketitle

\begin{abstract}
The widespread application of 3D human pose estimation (HPE) is limited by resource-constrained edge devices like Jetson Nano, requiring more efficient models.
A key approach to enhancing efficiency involves designing networks based on the structural characteristics of input data. However, effectively utilizing the structural priors in human skeletal inputs remains challenging.
To address this, we leverage both explicit and implicit spatio-temporal priors of the human body through innovative model design and a pre-training proxy task.
First, we propose a Nano Human Topology Network (NanoHTNet), a tiny 3D HPE network with stacked Hierarchical Mixers to capture explicit features.
Specifically, the spatial Hierarchical Mixer efficiently learns the human physical topology across multiple semantic levels, while the temporal Hierarchical Mixer with discrete cosine transform and low-pass filtering captures local instantaneous movements and global action coherence.
Moreover, Efficient Temporal-Spatial Tokenization (ETST) is introduced to enhance spatio-temporal interaction and reduce computational complexity significantly.
Second, PoseCLR is proposed as a general pre-training method based on contrastive learning for 3D HPE, aimed at extracting implicit representations of human topology.
By aligning 2D poses from diverse viewpoints in the proxy task, PoseCLR aids 3D HPE encoders like NanoHTNet in more effectively capturing the high-dimensional features of the human body, leading to further performance improvements.
Extensive experiments verify that NanoHTNet with PoseCLR outperforms other state-of-the-art methods in efficiency, making it ideal for deployment on edge devices like the Jetson Nano.
Code and models are available at \href{https://github.com/vefalun/NanoHTNet}{\textit{https://github.com/vefalun/NanoHTNet}}.

\end{abstract}

\begin{IEEEkeywords}
3D Human Pose Estimation, Human Topology, Efficient, Contrastive Learning.
\end{IEEEkeywords}

\IEEEpeerreviewmaketitle
\section{Introduction}
\IEEEPARstart{3}{D} human pose estimation (HPE) is a fundamental computer vision task aimed at estimating 3D human joint positions from images or videos. 
Recently, many effective 3D HPE methods~\cite{poseformer, mhformer,p-stmo, STCFormer, PoseFormerV2} have been proposed and widely applied in human-computer interaction contexts such as health monitoring~\cite{health}, rehabilitation training~\cite{Rehabilitation}, and virtual reality~\cite{vr}.
Notably, the deployment of these methods on edge devices like wearables, smart cameras, and mobile phones has significantly expanded the potential applications of 3D vision technologies.

However, deploying 3D HPE methods on these edge devices presents significant efficiency challenges. 
The mainstream approaches in 3D HPE~\cite{simplebaseline,videopose,lin2019trajectory, 2025_1, 2025_2} typically adopt a two-stage pipeline. 
In the first stage, lightweight 2D HPE models~\cite{chen2018cascaded,hrnet,rtmpose} are employed to estimate 2D poses from images or videos in real time. 
In the second stage, these 2D poses are subsequently lifted to 3D poses. 
However, as illustrated in Fig.~\ref{fig:motivation}, 
these methods fail to achieve real-time inference in the second stage when deployed on the Jetson Nano, a widely used AI edge device.
One key factor contributing to this limitation is that these 2D-to-3D pose lifting methods rely on pixel-based vision approaches that overlook the spatio-temporal priors inherent in human topology, resulting in redundant computations.
This leads us to ask:

 \begin{figure}[t]
    \centering
    \includegraphics[width=0.97\linewidth]{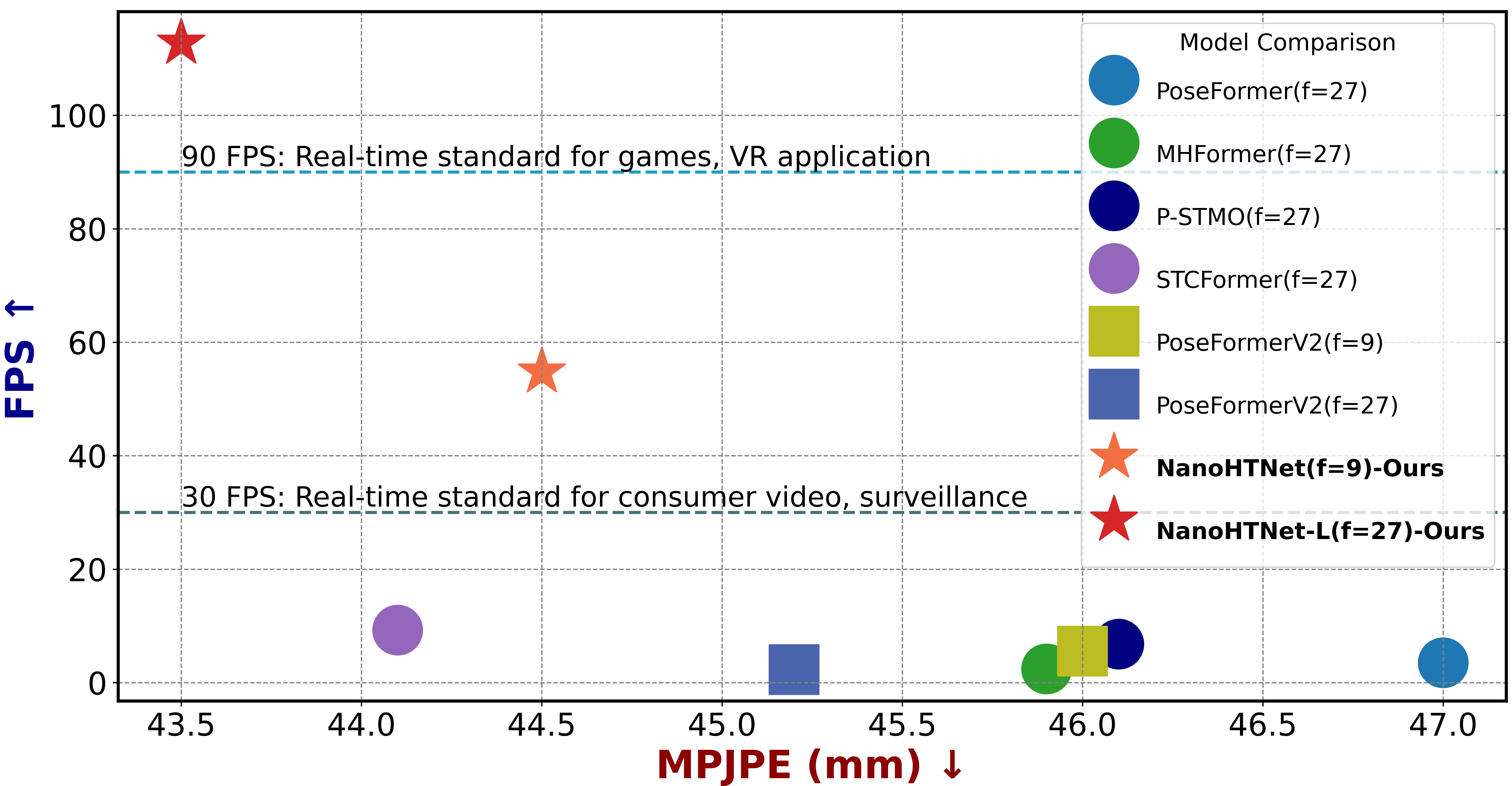}
    \VspaceFa
    \caption
    {
        Comparison of performance (MPJPE) and speed (FPS) between the proposed NanoHTNet and SOTA methods on the Jetson Nano, a compact and power-efficient AI device.
        NanoHTNet significantly excels in the speed-accuracy trade-off,  enabling real-time inference for 3D HPE in edge AI.
    }
    \VspaceFa
    \label{fig:motivation}
\end{figure}

\begin{itemize}
\item[$\bullet$]  \textit{How to better leverage spatio-temporal human priors to enhance data comprehension for more efficient 3D HPE? }
\end{itemize}

To address this, we explore human priors from the following three perspectives:

 \begin{figure}[t]
    \centering
    \includegraphics[width=0.95\linewidth]{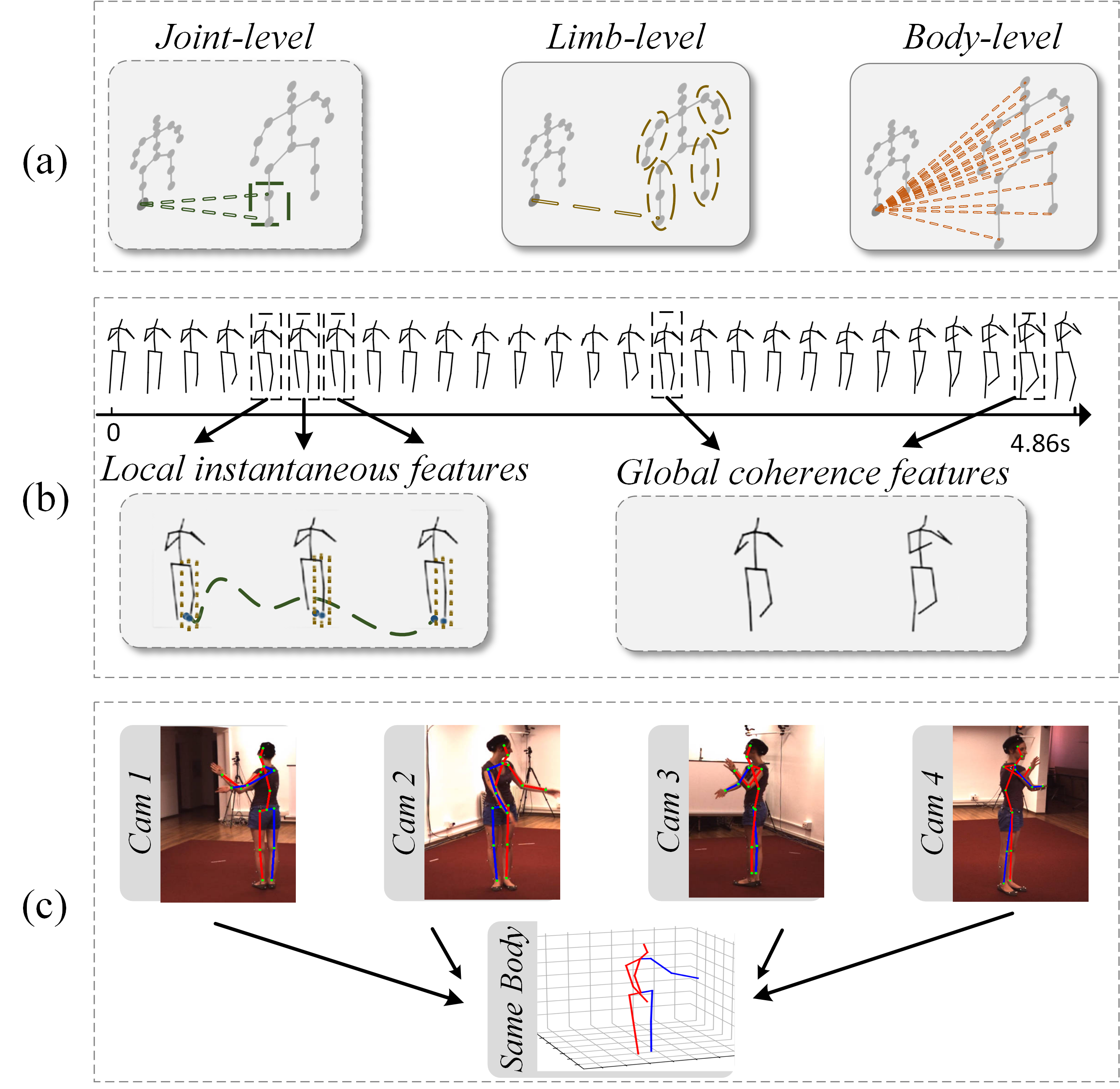}
    \caption
    {Spatio-temporal priors of the human skeleton:
        (a) Spatial hierarchical structure;
        (b) Temporal multi-scale kinematics;
        (c) Cross-view spatio-temporal topological consistency.
    }
    \VspaceFa
    \label{fig:mov_4}
\end{figure}

\VspaceSen
\textit{\textbf{(i)}} \textbf{\textit{Spatial hierarchical structure}}: 
The human body can be anatomically divided into multiple hierarchical levels: joints, parts, and the whole body (see in Fig.\figtwo{fig:mov_4}{a}).
At the joint level, joints are connected locally through the skeleton; 
at the part level, the joints in the same part cooperate closely during movement, and they will always advance and retreat together; 
at the body level, the inter-part joints are also globally connected with each other. 
For instance, running, a fundamental human activity, is characterized by the natural coordination of contralateral limbs, where the left arm synchronizes with the right leg, rather than ipsilateral limb movement.

\VspaceSen
\textit{\textbf{(ii)}} \textbf{\textit{Temporal multi-scale kinematics}}: Given the kinematic principle, the temporal movement dynamics of human can be divided into local instantaneous features and global coherence features. As shown in Fig.\figtwo{fig:mov_4}{b}, 
local features mainly focus on the immediate movements between adjacent frames, including variations in joint angles, velocity, and acceleration, which are crucial for predicting short-term movements.
Meanwhile, global features within the sequence (\textit{e.g.}, a 243-frame input spanning 4.86 seconds) concentrate on the overall coherence and coordination of actions. 
This enables the comprehension of global movement patterns and rhythms, such as the repetitive and cyclical nature in running, thereby aiding the model in better understanding the consistency of human motion.

\VspaceSen
\textit{\textbf{(iii)}} \textbf{\textit{Cross-view spatio-temporal topological consistency}}: The topological architecture of the human body exhibits consistent spatial representation across all viewpoints at the same moment as shown in Fig.\figtwo{fig:mov_4}{c}.  
Such consistency is crucial for developing robust models that are capable of generalizing across a wide range of poses and orientations. 
In insight of this, models can more accurately predict joint coordinates by aligning body parts from various perspectives.

Driven by the first two observations, a Nano Human Topology Network (NanoHTNet) is proposed.
NanoHTNet utilizes a dual-stream architecture to parallelly extract spatial and temporal kinematic features of human topology through stacked Hierarchical Mixers.
Specifically, the spatial Hierarchical Mixers, composed of Local Joint-level Connection (LJC), Intra-Part Constraint (IPC), and Global Body-level Interaction (GBI) modules, capture the hierarchical features of the human skeleton from joints to the entire body.
The temporal Hierarchical Mixers, with Instantaneous Motion Extraction (IME) and Global Coherence Perception (GCP) modules, learn both short-term and long-term motion characteristics.
\added{Moreover, Discrete Cosine Transform (DCT) separates low-frequency global kinematics from high-frequency noise. 
A Low-pass Filter (LF) is then applied to remove this noise and reduce data redundancy.} 
To further enhance spatio-temporal feature interaction, Efficient Temporal-Spatial Tokenization (ETST) module is proposed. 
\added{ETST rearchitects data mapping by creating fused tokens: temporal tokens that encapsulate the full body joints per frame and spatial tokens that represent each joint's entire trajectory. 
This strategy strengthens feature interaction from the outset while significantly reducing complexity and parameters.}

Further inspired by the cross-view topological consistency, we introduce PoseCLR, a self-supervised learning method utilizing contrastive learning to extract structural priors for better model parameter initialization.
Specifically, in the pre-training stage, a contrastive proxy task is designed to align the 2D poses of the same person at the same moment across different viewpoints, effectively capturing high-dimensional feature representations of the human body. 
In the fine-tuning phase, the pre-trained parameters are loaded into the encoder to train for the monocular 3D human pose estimation task. 
The encoder could be various mainstream 2D-to-3D lifting implementations, including NanoHTNet, GraFormer~\cite{GraFormer}, MixSTE~\cite{mixste}, PoseFormerV2~\cite{PoseFormerV2}, etc. 

Therefore, by leveraging unlabeled multi-view data or the inherent viewpoint information in existing datasets~\cite{ionescu2013human3}, 
PoseCLR could introduce better initialization parameters that represent the implicit structural prior of human topology for 3D HPE encoders.
This approach achieves a generic pre-training approach for 2D-to-3D lifting HPE without extra labeled data and model size increase.

In summary, the main contributions are as follows: 
\begin{itemize}[itemsep=4pt,topsep=0pt,parsep=0pt]

\item A Nano Human Topology Network (NanoHTNet) is proposed, which can infer on edge AI devices like Jetson Nano with excellent efficiency.

\item Spatial and temporal Hierarchical Mixers with a channel-split progressive structure are designed to efficiently learn multilevel spatio-temporal features of the human body.

\item \added{To the best of our knowledge, PoseCLR is the first to introduce a multi-view contrastive learning framework to the 2D-to-3D HPE task}, which utilizes cross-view consistency of the human body to optimize model initialization.

\item Experiments show that NanoHTNet effectively extracts explicit spatio-temporal human features, while PoseCLR, as a general pre-training method for 2D-to-3D HPE, further captures implicit features for better comprehension of the human body, leading to efficient 3D HPE.
\end{itemize}

\VspaceSen
This paper is the extension of our conference paper~\cite{htnet}, and NanoHTNet extends in the following aspects:
\added{(1) We introduce a Temporal Hierarchical Mixer augmented with DCT and LF to efficiently model multi-scale motion from video, a capability largely absent in the original spatial-focused work.
(2) We propose a fundamentally lighter architecture, NanoHTNet, which utilizes a novel Efficient Temporal-Spatial Tokenization (ETST) module to drastically reduce model complexity for real-time edge deployment.
(3) We introduce PoseCLR, a new contrastive learning pre-training method that captures implicit structural priors from multi-view data, a paradigm not present in the conference version.}

\section{Related Work}
\subsection{Efficient 3D HPE}

The deployment of real-time and accurate 3D HPE algorithms on edge devices could significantly promote the widespread application of 3D human vision technology. 
However, the hardware limitations of these devices impose constraints on model size and computational complexity.
To address this challenge, many methods try to enhance efficiency by data optimization.
DeciWatch~\cite{deciwatch} introduces a sample-denoise-recover pipeline, achieving 10× efficiency improvement on edge devices by estimating from less than sampled 10\% of video frames. 
HoT~\cite{hot}, as a plug-and-play framework, utilizes clustering algorithms to select a few representative tokens from redundant video sequences.
\added{PoseFormerV2~\cite{PoseFormerV2} explores the frequency domain by using a few low-frequency Discrete Cosine Transform (DCT) coefficients to create a compact representation of long skeleton sequences. This approach fuses global frequency-domain features with time-domain features from a few central frames within a specialized Time-Frequency Feature Fusion module. Our NanoHTNet also utilizes DCT and low-pass filtering for data optimization. However, instead of a fusion module, we employ a dual-stream architecture where the filtered frequency features are processed by a dedicated temporal mixer in parallel with a spatial mixer, allowing for specialized extraction of kinematic and physical topology.}

Other approaches for efficient HPE are based on the model lightweight. 
MobileHumanPose~\cite{mobilehumanpose} significantly reduces model parameters and FLOPs via skip cascading and residuals, facilitating deployment on mobile devices like Galaxy S20. 
MixSynthFormer~\cite{mixsynthformer} incorporates a reduction factor in an MLP-based mixed synthetic attention matrix generator, enabling dynamic fusion of temporal and spatial features and achieving real-time inference on an 8-core CPU Apple M1 Pro.
\added{Similarly, DSPNet ~\cite{dspnet} proposes a deep supervision pyramid network, which utilizes a lightweight up-sampling unit and a novel pyramid architecture to reduce computational load, achieving a remarkable balance between accuracy and efficiency.}
Our NanoHTNet integrates these methodologies by (1) incorporating DCT in data optimization to filter key features and (2) employing a channel-split progressive architecture and ETST module in model lightweight to reduce model complexity, resulting in  real-time inference on Jetson Nano.

\subsection{Human Topology Modeling in 3D HPE}
~\label{sec:2.2}
Unlike pixel-based vision tasks, 2D-3D lifting HPE takes human topology composed of sparse and highly structured joints as input, which integrates both the spatial, physical structures and temporal kinematic relationships~\cite{du2024kinematics, TIP2, TIP3}. 
Exploiting these inherent priors could boost estimation efficiency.

Many recent works~\cite{zeng2020srnet,wang2019not,wu2021limb} incorporate spatial physical structures from different body levels into the model design.
Some studies~\cite{SemGCN, stgcn, glagcn} utilize Graph Convolutional Neural Networks (GCNs) to model joint-level features, employing adjacency matrices to simulate the anatomical connections of joints and aggregate local features.
Addressing limb movement inconsistency, other works~\cite{zeng2020srnet, wang2019not, wu2021limb} extract features at the limb level.
For example, SRNet~\cite{zeng2020srnet} splits the joints of the human body into local groups and recombines them. 
Wang \textit{et al.}~\cite{wang2019not} treat body joints unequally and capture the relationships among the body parts with different DOFs. 
Wu \textit{et al.}~\cite{wu2021limb} design a limb pose-aware framework through angular relations to provide constraints for the limbs. 
Furthermore, vision Transformer~\cite{vit, transformer} enable encoding body-level dependencies via self-attention. This approach ~\cite{MotionBERT,sun2023mixsynthformer} captures interactions between the whole body, enhancing expressiveness in skeleton sequences. 
\added{More recently, diffusion models~\cite{diffpose1,diffpose2,finepose} have been used to tackle pose ambiguity by generating multiple topological hypotheses through a denoising process.}

Another line of works~\cite{tekin2016direct, mixste, Ktpformer} investigates temporal kinematics for better estimation.
Tekin \textit{et al.}~\cite{tekin2016direct} pioneers the integration of video sequences into 3D HPE based on the Seq2Frame pipeline, which only regresses the center frame, effectively leveraging motion information to resolve ambiguities.
MixSTE~\cite{mixste} introduces the Seq2Seq framework to the Transformer architecture for 3D HPE. 
It alternately extracts temporal and spatial relationships and simultaneously outputs 3D poses for all frames, enhancing motion consistency.
KTPFormer~\cite{Ktpformer} constructs kinematic topology and motion trajectory separately, which effectively enhances the global correlation modeling capability of human topology.
However, these methods can not simultaneously take full advantage of the spatio-temporal priors of human topology, leading to specific estimation errors and significant computational redundancy.

\subsection{Self-Supervised Learning in 3D HPE}
Beyond the inherent spatio-temporal priors of human topology, self-supervised learning can further capture high-dimensional and implicit representations of the human skeleton via unlabeled pre-training.
Current mainstream self-supervised learning methods can be categorized into generative proxy tasks~\cite{mae, mae2, mae3} and contrastive proxy tasks~\cite{simclr, moco, barlow, peCLR}. 
The former masks a portion of the input sequence and trains the model to predict the masked content. 
For example, P-STMO~\cite{p-stmo} employs masked pose modeling to pre-train the model by reconstructing randomly masked joints, thus enhancing the understanding of human body structures. 
MotionBERT~\cite{MotionBERT} extends the mask autoencoders framework to human motion representations, utilizing self-supervised pre-training to capture geometric, kinematic, and physical knowledge for multiple downstream tasks.

Contrastive learning captures the intrinsic characteristics of data by learning the relative relationships among samples within the dataset, providing a comprehensive understanding of the human body for skeleton-based tasks~\cite{crossclr, aimclr}. 
For example, UniHPE~\cite{unihpe} aligns images, corresponding 2D poses, and 3D poses as mutual positive samples for training, facilitating constraints and information sharing across different modalities. 
ORC-Pose~\cite{orcpose} uses unoccluded 2D poses as positive samples and occluded ones as negatives to ensure topology invariance, effectively mitigating the occlusion problem.
However, these methods overlook the multi-view data in datasets and the cross-view consistent topological patterns of the human topology. 
In this work, PoseCLR is designed to exploit these relative topological relationships to build contrastive pre-training tasks, thereby enhancing the extraction of topological priors.

\begin{figure*}[t]
    \centering
    \includegraphics[width=\linewidth]{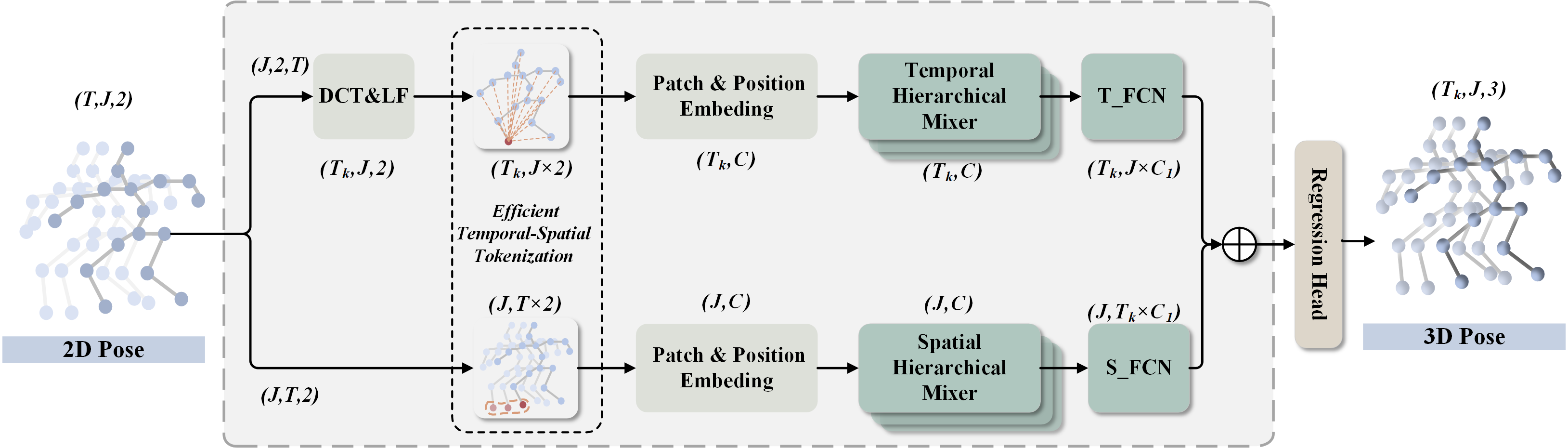}

    \caption
    {
        Overview of our encoding network NanoHTNet. A two-stream structure is designed for independently extracting the spatial-domain physical topology and the temporal-domain kinematic topology of the human body. \added{$T$ denotes input frames, while $T_k$ represents the reduced frames after applying DCT and LF. $C$ is the main channel dimension after patch embedding, and $C_l$ is the channel dimension after MLP networks $T\_FCN$ and $S\_FCN$}.
    }
    \VspaceFb
    \label{fig:overview}
\end{figure*}

\section{Methods}
Our method consists of two stages, each stage focusing on the extraction of topological features. 
In the pre-training phase, we implement a contrastive proxy task based on cross-view topological consistency to extract implicit and high-dimensional features without 3D labels;
In the fine-tuning phase, we design the encoder architecture to extract explicit hierarchical representations of human topology. Initialized with pre-trained parameters, the model is then fine-tuned using 3D ground truth supervision.
In Sec.~\ref{sec:Encoding}, we introduce the overview of NanoHTNet as the encoder.
Subsequently, Hierarchical Mixers, the core elements of NanoHTNet, are detailed in Sec.~\ref{sec:Mixers}. 
Finally, the contrastive learning-based method PoseCLR is presented in Sec.~\ref{sec:PoseCLR}.

\subsection{Encoding Network Structure}
\label{sec:Encoding}
As the encoder for the 3D HPE task, the overall structure of NanoHTNet is shown in Fig.~\ref{fig:overview}. 
Given a 2D pose sequence $X\in \mathbb{R}^{T\times J\times 2}$ with $T$ frames and $J$ joints, NanoHTNet adopts a tiny dual-stream architecture to extract and fuse spatial and temporal features in parallel.

\VspaceSen
\VspaceSen
\textbf{Spatial-domain.}
Referencing Transformer-based methods, $X\in \mathbb{R}^{T\times J\times 2}$ is initially reshaped to $X_s\in \mathbb{R}^{J\times (T\cdot2)}$, allowing each joint to be treated as an individual token. 
These tokens are mapped to a high dimension ${X_s}' \in \mathbb{R}^{J\times (T \cdot C)}$ with a linear layer in the patch embedding, where $C$ is the channel dimension. 
Then, we embed ${X}'$ with a learnable positional matrix $E_{pos} \in \mathbb{R}^{1\times (T \cdot C)}$ in the position embedding to obtain the embedded features $X^l$, where $l$ is the index of spatial Hierarchical Mixers.
These Hierarchical Mixers, which are described in detail in Sec.~\ref{sec:Mixers}, play a crucial role in the extraction of the human topology.
Finally, a channel MLP~\cite{mlpmixer} called S\_FCN is adopted to aggregate information between different channels.

\VspaceSen
\VspaceSen
\textbf{Temporal-domain.}
Long sequences, such as 351 frames~\cite{mhformer,strided} and 243 frames~\cite{STCFormer, MotionBERT}, enhance performance by expanding the receptive field. 
However, these extended sequences introduce substantial redundancy and lead to a quadratic increase in computational complexity.
To mitigate video input redundancy while maintaining a large receptive field, we incorporate a Discrete Cosine Transform (DCT) and a low-pass filter before the patch embedding for data compression and noise reduction. First, we utilize DCT to transform the input data from the time domain to the frequency domain as:
\begin{equation}
\label{equ1}
\small
X_{\phi}=c(f)\sum_{t=0}^{T-1} X_t \cos \left[ \frac{\pi(2T+1)f}{2T}\right], f=0,1,...,T-1,
\end{equation}
where ${X_t}' \in \mathbb{R}^{J\times C \times T}$ is the input with $T$ frames, $f$ represents the index of the frequency component, and $c(f)$ is a normalization coefficient used to ensure the orthogonality and energy conservation of DCT.
Then, we utilize low-pass filtering  to reduce redundancy as:
\begin{equation}
\label{equ2}
X{^k}_{\phi} = \begin{cases} 
X_{\phi}, & \text{if } f < k \\
0, & \text{otherwise},
\end{cases}
\end{equation}
where $k$ is the cutoff frequency, and $X{^k}_{\phi}$ represents the frequency domain feature after filtering.
Low-frequency coefficients in the frequency domain focus primarily on global movement trends and major poses, while high-frequency coefficients contain noise and minor variations. 
Through low-pass filtering, our model focuses on the main movements and poses, boosting computational efficiency and generalization within a large receptive field.
The subsequent steps mirror those in the spatial domain, but each frequency domain feature is treated as a token.
These tokens are processed through the Temporal Hierarchical Mixer and T\_FCN. 

Finally, we concatenate the spatio-temporal representations in the dual-stream network and obtain the 3D pose $Y \in \mathbb{R}^{T_k\times J \times 3}$ through a regression head.

\begin{figure*}[t]
    \centering
    \includegraphics[width=0.95\linewidth]{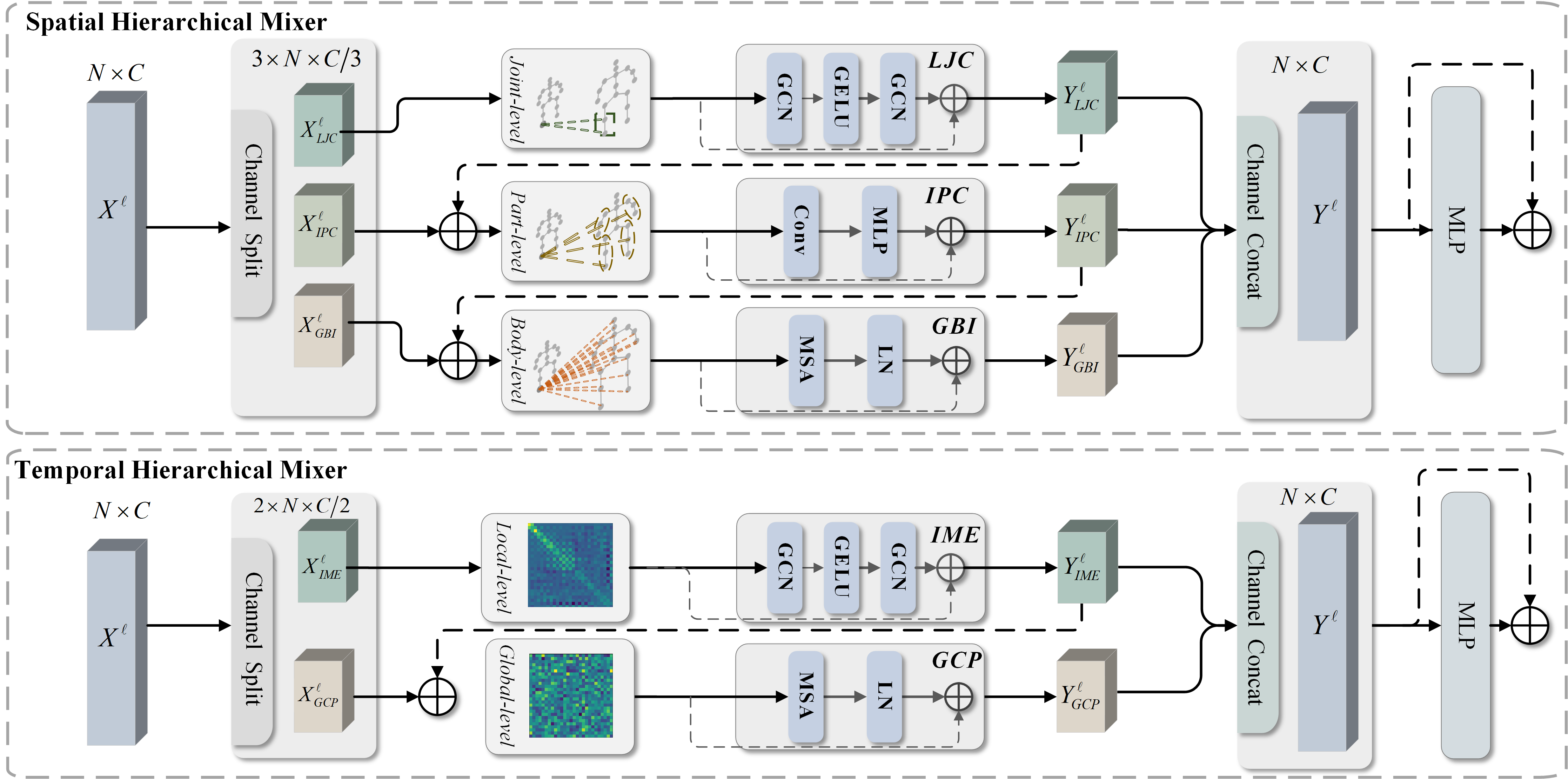}

    \caption
    {   Stacked Hierarchical Mixers are key in our encoder. 
        Each Spatial Hierarchical Mixer consists of Local Joint-level Connection (LJC), Intra-Part Constraint (IPC), and Global Body-level Interaction (GBI).
        Each Temporal Hierarchical Mixer effectively captures temporal features from local to global extents by Instantaneous Motion Extraction (IME) and Global Coherence Perception (GCP).
    }
    \VspaceFb
    \label{fig:mixers}
\end{figure*}

\VspaceSen
\textbf{Efficient Temporal-Spatial Tokenization.}
Existing Video Pose Transformers~\cite{poseformer, mhformer, MotionBERT, mixste, STCFormer} construct tokens independently within spatial and temporal dimensions (Spatially, $X\in \mathbb{R}^{T\times J\times 2} \rightarrow {X_s}' \in \mathbb{R}^{T\times J \times C)}$; Temporally, $X\in \mathbb{R}^{T\times J\times 2}  \rightarrow {X_t}' \in \mathbb{R}^{J\times T \times C}$). 
These approaches, we termed Decoupled Temporal-Spatio Tokenization (DTST), fail to capture the interaction of spatio-temporal information, and the computational complexity increases quadratically with the number of tokens as shown in Tab.~\ref{tab:complexity_comparison_revised}.

To construct a tiny and more efficient network, we introduce Efficient Temporal-Spatial Tokenization (ETST).
Specifically, in the temporal processing stream, $X \in \mathbb{R}^{T \times J \times 2}$ is reshaped to $X_t \in \mathbb{R}^{T \times (J \cdot 2)}$ and subsequently embedded to ${X_t}' \in \mathbb{R}^{T \times C}$. 
This enables the model to track the global motion trends of all joints across frames, enhancing spatio-temporal interactions and the understanding of human motion.
Similarly, in the spatial stream, $X \in \mathbb{R}^{T \times J \times 2}$ is reshaped to $X_s \in \mathbb{R}^{J \times (T \cdot 2)}$ and then embedded to ${X_s}' \in \mathbb{R}^{J \times C}$, allowing the model to focus on spatial interactions between joints over $T$ frames.

ETST facilitates the fusion of spatio-temporal information while effectively reducing computational complexity from $O(J^2 \times T \times C) + O(T^2 \times J \times C)$ to $O(J^2 \times C) + O(T^2 \times C)$. 
By integrating the Low-pass Filter, the computational complexity can be further reduced to $O(J^2 \times C) + O(T_k^2 \times C)$. 
As the number of temporal input frames increases (like PoseFomer~\cite{poseformer} with 243 frames and StridedFormer~\cite{strided} with 351 frames), the efficiency benefits become more apparent, making it suitable for deployment on resource-constrained edge devices like the Jetson Nano.

\begin{table}[t]
\renewcommand{\arraystretch}{1.5}
\caption{\added{Complexity comparison between different Tokenizations}}
\label{tab:complexity_comparison_revised}
\centering
\begin{tabular*}{0.9\columnwidth}{@{\extracolsep{\fill}}llcc}
\hline
\textbf{\added{Method}} & \textbf{\added{Domain}} & \textbf{\added{Tokens}} & \textbf{\added{Complexity}} \\
\hline
\multirow{2}{*}{\added{DTST}}  & \added{Temporal} & \added{T}    & \added{$ O(J \times T^2 \times C)$} \\
& \added{Spatial}  & \added{J}       & \added{$ O(T \times J^2 \times C)$} \\
\hline
\multirow{2}{*}{\added{ETST (Ours)}} & \added{Temporal} & \added{$T_k$}   & \added{$O(T_k^2 \times C)$} \\
& \added{Spatial}  & \added{J}       & \added{$O(J^2 \times C)$} \\
\hline
\end{tabular*}
\end{table}
\vspace{-2mm}

\begin{figure}[t]
    \centering
    \includegraphics[width=1.0\linewidth]{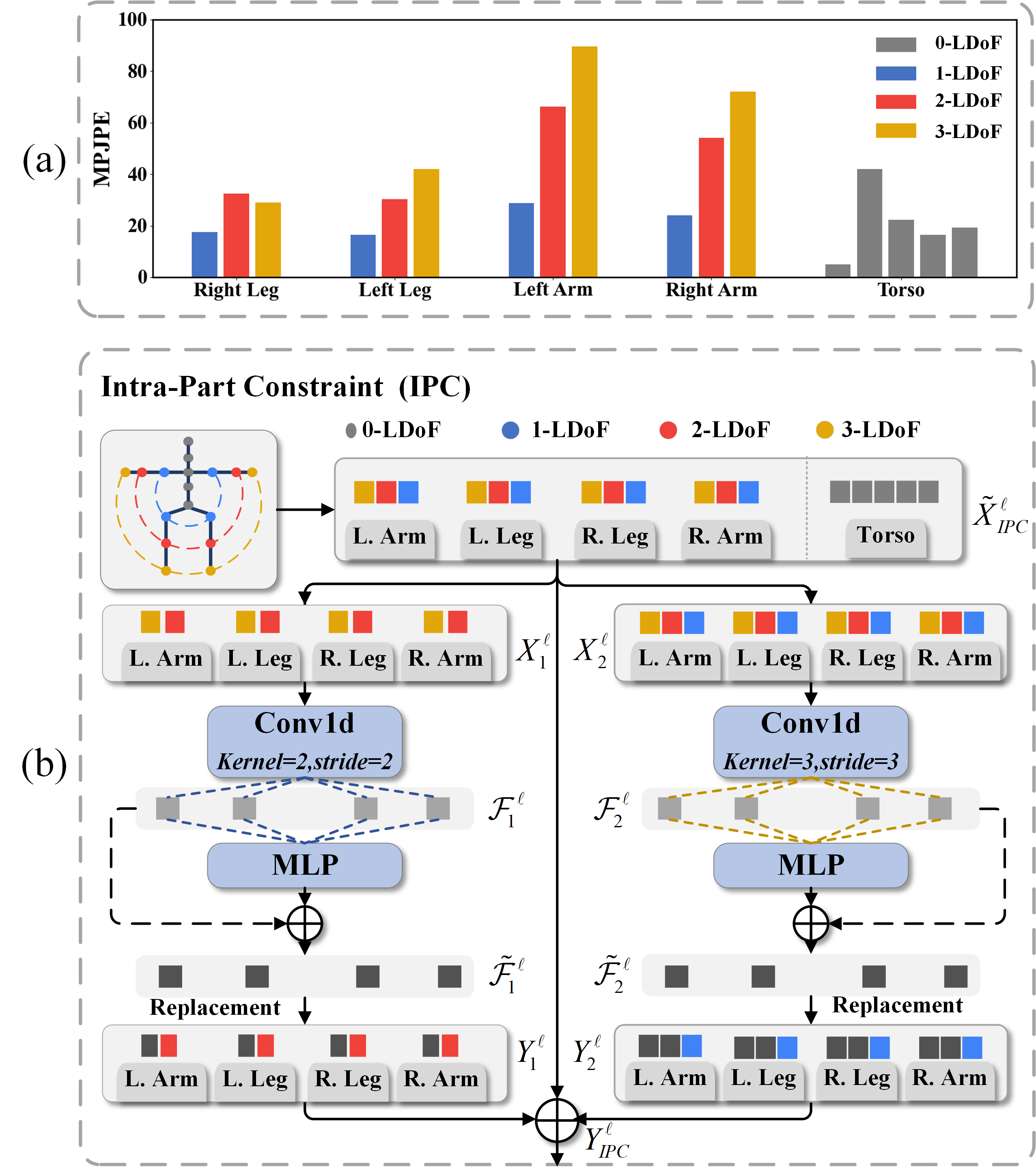}
    \caption
    {
        (a) Specific estimation error distribution caused by neglect of part-level features; (b) Structure of Intra-Part Constraint (IPC) module.
    }
    \VspaceFb
    \label{fig:ipc}
\end{figure}

\subsection{Hierarchical Mixers}
\label{sec:Mixers}
The core of our encoder is composed of stacked Hierarchical Mixers. We design  spatial Hierarchical Mixers and temporal Hierarchical Mixers to capture the spatial anatomical topology and temporal kinematic topology, respectively.

\begin{figure*}[t]
    \centering
    \includegraphics[width=0.95\linewidth]{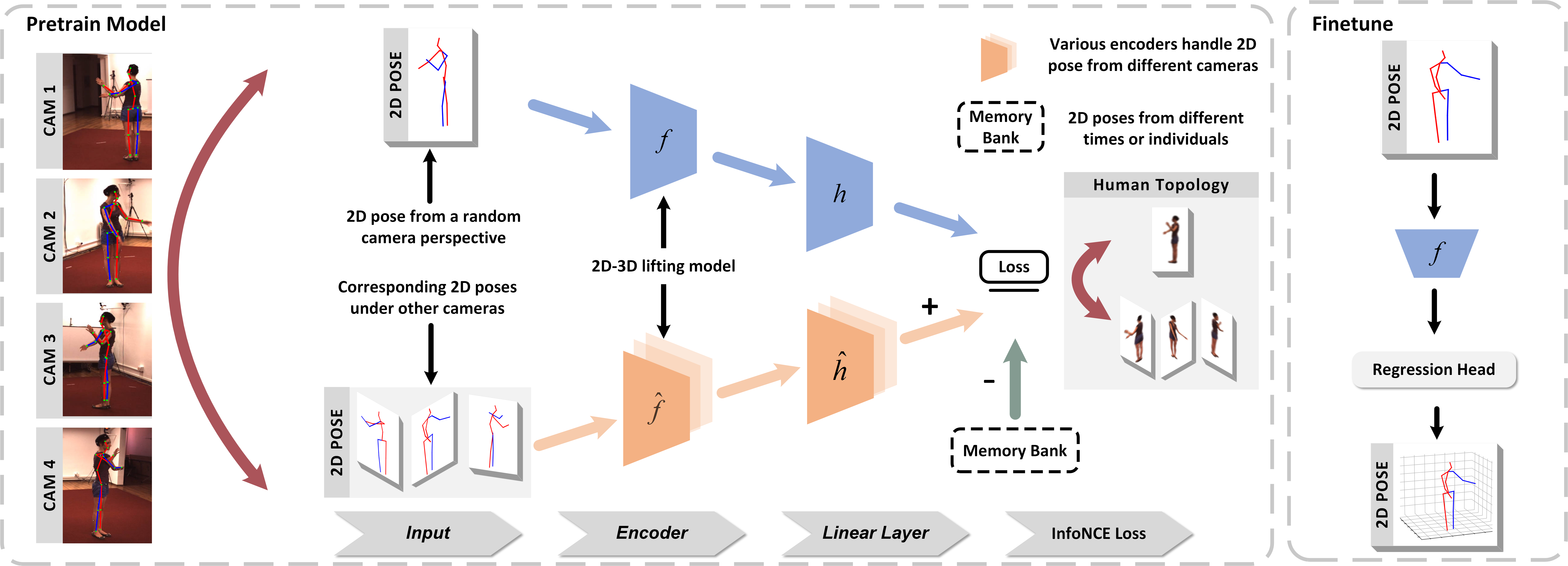}

    \caption
    {
        PoseCLR is a contrastive learning architecture based on cross-view topological consistency. 
        2D poses of the same individual captured simultaneously from different cameras are treated as positive pairs, while other 2D poses as negative pairs.
    }
    \VspaceFb
    \label{fig:poseclr}
\end{figure*}

\textbf{Spatial Hierarchical Mixers.}
To learn hierarchical human topology representations, the input $X_s$ is split into three parts: $X^s_{\mbox{\textit{\tiny LJC}}}$, $X^s_{\mbox{\textit{\tiny IPC}}}$, and $X^s_{\mbox{\textit{\tiny GBI}}}$, which have equal dimensions $C'=C/3$ and are fed into corresponding modules: Local Joint-level Connection (LJC), Intra-Part Constraint (IPC), and Global Body-level Interaction (GBI) as shown in Fig.~\ref{fig:mixers}.
These modules construct the human topology from multiple semantic levels: joint, part, and body.

\textit{1) Local Joint-level Connection.}
LJC mainly focuses on joint-level features by leveraging GCNs~\cite{mgcn} to model the natural relations among joints locally. The graph-structured human skeleton can be defined as $G = (V,A)$, where $V$ is a set of $J$ joints and $A \in \left \{ 0,1 \right \}^{J\times J }$ is an adjacency matrix representing the connection between joints. 
Given the input $X^{s}_{\mbox{\textit{\tiny LJC}}} \in \mathbb{R}^{J\times C'}$,
LJC aggregates neighboring joint features into $Y^{s}_{\mbox{\textit{\tiny LJC}}}$, given by:
\begin{equation}
\label{eq3}
\begin{array}{l}
\mbox{\textit{GCN}}(X^s_{\mbox{\textit{\tiny LJC}}}) = \tilde{D}^{-\frac{1}{2}}\tilde{A} \tilde{D}^{-\frac{1}{2}}X^{s}_{\mbox{\textit{\tiny LJC}}}W, \\
\vspace{2mm}
Y^{s}_{\mbox{\textit{\tiny LJC}}} = X^{s}_{\mbox{\textit{\tiny LJC}}} + \mbox{\textit{GCN}}(\sigma(\mbox{\textit{GCN}}(X^{s}_{\mbox{\textit{\tiny LJC}}}))) ,
\end{array}
\end{equation}
where $\tilde{A}=A+I$, $\tilde{D}$ is the diagonal node degree matrix, $W$ is the weight matrix, and $\sigma(\cdot)$ denotes the $\mbox{\textit{GELU}}$ activation function.

\textit{2) Intra-Part Constraint.}
Neglecting part-level features of the human body often leads to special estimation errors.
Taking Fig.\figfive{fig:ipc}{a} as an example, the previous state-of-the-art model~\cite{hopfir} suffers from significant estimation errors on limb-end joints, and these errors are regular for joints with different Limb Degree of Freedoms (LDoFs), where LDoF denotes the distance between the joints and the torso, \textit{e.g.}, the wrist with 3-LDoF has a larger error than the elbow with 2-LDoF.
A novel IPC (see in Fig.\figfive{fig:ipc}{b}) is designed, which mainly focuses on the part-level features within each limb through topological constraints. Given the input $\tilde{X}^s_{\mbox{\textit{\tiny IPC}}} = X^s_{\mbox{\textit{\tiny IPC}}} + Y^s_{\mbox{\textit{\tiny LJC}}}$, topological constraints perform in two sets of joints: \textit{(\romannumeral1)} $X^s_{\textit{1}} \in {\mathbb{R}^{8 \times C'}}$ consists of end and 2-LDoF joints; \textit{(\romannumeral2)} $X^s_{\textit{2}} \in {\mathbb{R}^{12 \times C'}}$ consists of all limb joints. Next, $X^s_{\textit{j}},j \in \{1,2\}$ are fed into two convolution layers to generate limb features $\mathcal{F}^s_{j}$. Then, a channel MLP~\cite{mlpmixer} is adopted to aggregate information between different channels for each limb feature:
\begin{equation}
\setlength{\abovedisplayskip}{5pt}
    \begin{array}{l}
    \mathcal{F}^s_{j}=\sigma(\mbox{\textit{Conv}}_{j}(X^s_{j})), j\in\{1,2\},\\[0.1cm] 
    \mathcal{\tilde{F}}^s_{j}=\mathcal{F}^s_{j} + \mbox{\textit{MLP}}(\mbox{\textit{LN}}(\mathcal{F}^s_{j})),
    \end{array}
\setlength{\belowdisplayskip}{2pt}
\end{equation}
where $\mathcal{\tilde{F}}^s_{j}$ are aggregated limb features, $LN(\cdot)$ denotes Layer Normalization, $\mbox{\textit{Conv}}_{\textit{1}}$ and $\mbox{\textit{Conv}}_{\textit{2}}$ are convolution layers with kernel sizes of 2 and 3, respectively. 
Based on the above $\mathcal{\tilde{F}}^s_{\textit{j}}$, we construct two topological constraints $\mathcal{R}_{j}$ in $X^s_\textit{j}$: 
\textit{(\romannumeral1)} In $\mathcal{R}_{1}$, 3-LDoF joints in $X^s_\textit{1}$ are replaced with $\mathcal{\tilde{F}}^s_{\textit{1}}$;
\textit{(\romannumeral2)} In $\mathcal{R}_{2}$, 3-LDoF and 2-LDoF joints in $X^s_\textit{2}$ are replaced with $\mathcal{\tilde{F}}^s_{\textit{2}}$.
Then, the unprocessed joints are filled into the replaced $X^s_j$, and $Y^s_{\mbox{\textit{\tiny IPC}}}$, the final output of IPC, can be represented as follows:
\begin{equation}
\setlength{\abovedisplayskip}{5pt}
    Y^s_{\mbox{\textit{\tiny IPC}}} = \tilde{X}^s_{\mbox{\textit{\tiny IPC}}} + \mathcal{R}_{1}(\tilde{X}^s_{\mbox{\textit{\tiny IPC}}}, \mathcal{\tilde{F}}^s_{1}) + \mathcal{R}_{2}(\tilde{X}^s_{\mbox{\textit{\tiny IPC}}}, \mathcal{\tilde{F}}^s_{2}).
\setlength{\belowdisplayskip}{4pt}
\end{equation}
\added{The aggregated limb feature $\mathcal{\tilde{F}}^s_{j}$ provides a holistic and more stable representation of the limb's motion. The Replacement operation $\mathcal{R}_{j}$ uses this robust feature to constrain the error-prone distal joints, thereby severing the error propagation path. Finally, the residual connection allows the network to learn a corrective update based on this topological prior, leading to a more kinematically consistent and accurate pose estimation}.

\textit{3) Global Body-level Interaction.}
GBI is built by the self-attention of Transformer, which can capture global contexts across the whole human body~\cite{mhformer,GraFormer}.
Given $\tilde{X}^s_{\mbox{\textit{\tiny GBI}}} = X^s_{\mbox{\textit{\tiny GBI}}} + Y^s_{\mbox{\textit{\tiny IPC}}}$ as input, the GBI module can be formulated as:
\begin{equation}
\setlength{\abovedisplayskip}{3pt}
    \begin{array}{l}
    H_i=\mbox{\textit{Softmax}}({Q^{s}{K^{s}}^{T}}{/}{\sqrt{C'}})V^{s},i\in\{1,...,h\}\\[0.05cm]
        \mbox{\textit{MSA}}(\tilde{X}^s_{\mbox{\textit{\tiny GBI}}})=\mbox{Concat} (H_1,H_2,...,H_h)W_{out}, \\[0.05cm]
        \tilde{Y}^s_{\mbox{\textit{\tiny GBI}}} = \tilde{X}^s_{\mbox{\textit{\tiny GBI}}} + \mbox{\textit{LN}}(\mbox{\textit{MSA}}(\tilde{X}^s_{\mbox{\textit{\tiny GBI}}})),
    \end{array}
\setlength{\belowdisplayskip}{1pt}
\label{eq2}
\end{equation}
where $h$ is the number of attention heads, $Q^{s},K^{s},V^{s}$ are query, key, and value matrices, which are calculated from $\tilde{X}^s_{\mbox{\textit{\tiny GBI}}}$ by linear transformations, $\tilde{Y}^s_{\mbox{\textit{\tiny GBI}}}$ is the output of the GBI module.

\begin{table*}[ht]
\footnotesize
\centering
\caption
{ 
  Quantitative comparisons with video-based methods on Human3.6M under MPJPE and P-MPJPE. 
  Frames denote the number of frames input. RF is the effective Receptive Field of the input sequence.
  FLOPs (Floating Point Operations Per Second) quantify computational complexity.
  FPS (Frame Per Second) is implemented on Jetson Nano. 
}
\setlength{\tabcolsep}{3.5mm}
\renewcommand{\arraystretch}{1.3} 
\begin{tabular}{lc|ccccc|cc}
\toprule [1pt]
Method &Seq2Seq  &Frames &RF &Param (M) &FLOPs (M)  &FPS  &MPJPE$\downarrow$ &P-MPJPE$\downarrow$ \\

\midrule [0.5pt]
PoseFormer \cite{poseformer}   &\XSolidBrush &27 &27 &9.59 &452 &3.53 &47.0 &- \\
StridedFormer \cite{strided}  &\XSolidBrush &27 &27 &4.01 &128 &5.36 &46.9 &- \\
StridedFormer \cite{strided}  &\XSolidBrush &81 &81 &4.06 &392 &4.17 &45.9 &- \\
P-STMO \cite{p-stmo}   &\XSolidBrush    &27   &27 &4.60 &82 &7.04 &48.2 &-\\
P-STMO \cite{p-stmo}   &\XSolidBrush    &27   &243 &4.60 &82 &6.81 &46.1 &-\\
MHFormer \cite{mhformer}  &\XSolidBrush &27   &27 &18.92 &1031 &2.43 &45.9 &36.1 \\
MixSTE \cite{mixste}   &\Checkmark    &3   &3 &33.65 &3420 &5.67 &49.6 &38.9\\
PoseFormerV2\cite{PoseFormerV2}    &\XSolidBrush   &9 &81 &14.35 &352 &5.55 &46.0 &36.1\\
PoseFormerV2\cite{PoseFormerV2}  &\XSolidBrush  &27  &243 &32.35 &1055 &2.30 &45.2 &35.6\\
STCFormer\cite{STCFormer}   &\Checkmark  &27    &27 &4.75 &4346 &9.26 &44.1 &34.7\\
\added{MotionAGFormer-XS\cite{motionagformer}} &\added{\Checkmark}  &\added{27}    &\added{27} &\added{2.20} &\added{1937} &\added{19.2} &\added{45.1} &\added{36.9}\\
\midrule [0.5pt]
\rowcolor[HTML]{DADADA}
\added{HTNet-L (Conf. ver.) ~\cite{htnet}} &\added{\Checkmark}  &\added{27} &\added{27} &\added{10.1} &\added{631}  &\added{5.12} &\added{46.1} &\added{36.0}\\
\rowcolor[HTML]{DADADA}
NanoHTNet (Ours) &\Checkmark  &9 &243 &\textbf{1.52} &\textbf{33}  &54.81 &44.5 &35.2\\
\rowcolor[HTML]{DADADA}

NanoHTNet-L (Ours) &\Checkmark  &27 &243 &4.40 &156 &\textbf{112.76}  &\textbf{43.5} &\textbf{34.5}\\
\bottomrule [1pt]

\end{tabular}
\label{table:video}
\end{table*}

\textbf{Temporal Hierarchical Mixers.}
To effectively extract temporal multi-scale kinematic features, we design the temporal Hierarchical Mixers. This design includes two components: \textit{(i)} Instantaneous Motion Extraction (IME) module, based on graph convolutional networks, captures local temporal features such as acceleration and velocity between adjacent frames; \textit{(ii)} Global Coherence Perception (GCP) module, based on self-attention mechanisms, learns periodic changes and inherent coherence of motion, thereby fully leveraging the kinematic features within long-term sequences.

Significantly, these mixers adopt a channel-split progressive architecture rather than a simple serial or parallel structure for efficient feature extraction. 
Specifically, channels are split into several parts and allocated to distinct sub-modules (\textit{e.g.}, LJC, IPC, GBI, IME, and GCP). 
Meanwhile, the output from the previous sub-module is input into the subsequent module via residual-like connections, learning the human topological features from local to global. Finally, the output of each sub-module is concatenated and fed into an MLP network for aggregation.
To summarize, Hierarchical Mixers can aggregate features from different levels and obtain a more expressive representation, achieving a balance between feature extraction and computational efficiency.

\subsection{PoseCLR}
\label{sec:PoseCLR}

In addition to the above explicit extraction of kinematic and physical features of the human body, we observe that human topology exhibits consistency across multiple views simultaneously. Building on this insight, we propose PoseCLR, a multi-view contrastive learning framework for implicit extraction of human topological features  (see Fig.~\ref{fig:poseclr}). This self-supervised learning approach consists of pre-training and fine-tuning stages, extracting universal skeletal priors from unlabeled multi-view 2D poses for more accurate 3D HPE.

\VspaceSen
\textbf{Pre-training Stage.}
During the pre-training stage, we utilize a memory-augmented architecture~\cite{moco, crossclr} to align 2D poses from different camera views. 
The goal is to maximize the similarity between 2D poses of the same individual captured simultaneously from various camera angles, as these poses correspond to the same human topology in world coordinates. 
We take the Human3.6M dataset for illustration, which includes four camera viewpoints.
In each iteration, we select a 2D pose from a random viewpoint as the input into encoder $f$. 
We employ NanoHTNet in Sec.~\ref{sec:Encoding} as the encoder, which can be any 2D-to-3D lifting network, to encode the input into universal high-dimensional human representations $h$.
Meanwhile, the poses from all four viewpoints are treated as positive pairs and inputted into $\hat{f}$. 
Specifically, four $\hat{f}$ are the momentum updated version of $f$ for four distinct views, encoding these inputs into the topologic representations $\hat{h}$.
Subsequently, we employ the InfoNCE loss~\cite{infonce} to minimize the distance between $h$ and $\hat{s}$ representing the same topology, while maximizing the distance between features of different topologies:
\begin{equation}
\mathcal{L}_{p} = -\log \frac{\exp(\mathcal{S}(q, k^+)/\tau)}{\exp(\mathcal{S}(q, k^+)/\tau) + \sum_{i=0}^{K}\exp(\mathcal{S}(q, k_i^-)/\tau)},
\label{eq2}
\end{equation}
where $q$ represents the feature vector, $k^+$ denotes the positive samples, $k_i^-$ denotes the negative samples, $\mathcal{S}$ is the similarity function, $\tau$ is the temperature parameter (a tunable hyperparameter) used to control the scale of the similarity scores, and $K$ indicates the number of negative samples.
Notably, the positive samples are subsequently stored in a first-in-first-out memory bank, where they will serve as negative samples in future iterations.

\VspaceSen
\textbf{Finetuning Stage.}
The pre-training stage emphasizes generic human body representations, while the finetuning stage focuses on representations more aligned with the 3D HPE task. 
Unlike previous networks with randomly initialized parameters, we load encoder parameters $f$ from the pre-training stage, which benefits the network by: 
1) Improving performance. Pre-training enables the model to capture high-dimensional human topological prior, further assisting in the comprehension of human-based inputs;
2) Accelerating convergence. The model does not require learning from scratch but refines upon pre-learned features, which substantially reduces the training time.
In this phase, we introduce 3D pose labels and employ the simple Mean Per Joint Position Error (MPJPE) loss for supervised training:
\begin{equation}
   \mathcal{L}=\sum_{t=1}^{T} \sum_{j=1}^{J}\left\|Y_{j}^{t}-\tilde{X}_{j}^{t}\right\|_{2},
\end{equation}
where ${X}_{j}^{t}$ and $Y_{j}^{t}$ are the predicted and ground truth 3D poses of $j$ joint at $t$ frame.

\begin{table*}[t]
  \caption
  {
      Quantitative comparisons with image-based methods on Human3.6M under Protocol \#1 (MPJPE) and Protocol \#2 (P-MPJPE). 
      2D keypoints detected by CPN are used as inputs in the top and middle part of the table, and the ground truth of 2D poses are used as inputs in the bottom part of the table. 
  }
  \footnotesize
  \centering
  \renewcommand{\arraystretch}{1.25} 
  \resizebox{\textwidth}{!}{
  \begin{tabular}{@{}l|ccccccccccccccc|c}
  \toprule
  \textbf{Protocol \#1 (CPN)}  & Dir. & Disc & Eat & Greet & Phone & Photo & Pose & Purch. & Sit & SitD. & Smoke & Wait & WalkD. & Walk & WalkT. & Avg.\\
  \midrule

  Martinez \textit{et al.}.~\cite{simplebaseline} &51.8 &56.2 &58.1 &59.0 &69.5 &78.4 &55.2 &58.1 &74.0 &94.6 &62.3 &59.1 &65.1 &49.5 &52.4 &62.9 \\

  Fang \textit{et al.}~\cite{fang2018learning} & 50.1& 54.3& 57.0& 57.1& 66.6& 73.3& 53.4& 55.7& 72.8& 88.6& 60.3& 57.7& 62.7& 47.5& 50.6& 60.4 \\

  LCN ~\cite{ci2019optimizing} &46.8 &52.3& {44.7}& 50.4& 52.9& 68.9& 49.6& 46.4& 60.2 &78.9& 51.2& 50.0& 54.8& 40.4& 43.3& 52.7 \\

  ST-GCN ~\cite{stgcn}   &46.5 &48.8 &47.6& 50.9& 52.9 &61.3 &48.3 &45.8 &59.2 &64.4& 51.2& 48.4& 53.5& 39.2& 41.2& 50.6\\

    UGRN \cite{li2023pose} & 47.9 & 50.0 & 47.1 & 51.3 & 51.2 & 59.5 & 48.7 & 46.9 & {56.0} & {61.9} & 51.1 & 48.9 & 54.3 & 40.0 & 42.9 & 50.5 \\

  SRNet~\cite{zeng2020srnet}  &44.5& {48.2} &47.1 &{47.8} &51.2 &{56.8} &50.1& {45.6}& 59.9 &66.4 &52.1 &{45.3} &54.2 &39.1 &40.3 &49.9\\

  MLP-JCG \cite{tang2023mlp} & {43.8} & \textbf{46.7} & 46.9 & {48.9} & {50.3} & 60.1 & \textbf{45.7} & {43.9} & {56.0} & 73.7 & {48.9} &48.2 & {50.9} &39.9 &41.5 &49.7 \\

  \added{Gong \textit{et al.}~\cite{diffpose1}} &\added{42.8} &\added{49.1} &\added{45.2} &\added{48.7} &\added{52.1} &\added{63.5} &\added{46.3} &\added{45.2} &\added{58.6} &\added{66.3} &\added{50.4} &\added{47.6} &\added{52.0} &\added{37.6} &\added{40.2} &\added{49.7} \\

\added{PGFormer~\cite{pgformer}} &\added{43.8} &\added{49.9} &\added{45.5} &\added{49.0} &\added{51.0} &\added{58.2} &\added{47.0} &\added{45.7} &\added{58.1} &\added{65.8} &\added{49.6} &\added{46.5} &\added{53.6} &\added{38.0} &\added{40.6} &\added{49.5} \\

  MGCN~\cite{mgcn}  &45.4 &49.2 &45.7 &49.4 &{50.4} &58.2 &47.9 &46.0 &57.5 &63.0 &49.7 &46.6 &52.2 &38.9 &40.8 &49.4 \\

  RS-Net \cite{hassan2023regular} &44.7 &48.4 &44.8 &49.7 &49.6 &58.2 &47.4 &44.8 &55.2 &59.7 &49.3 &46.4 &51.4 &38.6 &40.6 &48.6 \\

  HopFIR~\cite{hopfir} &43.9 &47.6 &45.5 &48.9 &{50.1} &58.0 &46.2 &44.5 &55.7 &62.9 &49.0 &45.8 &51.8 &38.0 &39.9 &48.5 \\
  
  \added{GraphMLP~\cite{graphmlp}} &\added{43.7} &\added{49.3} &\added{45.5} &\added{47.9} &\added{50.5} &\added{56.0} &\added{46.3} &\added{44.1} &\added{55.9} &\added{\textbf{59.0}} &\added{48.4} &\added{45.7} &\added{51.2} &\added{\textbf{37.1}} &\added{39.1} &\added{48.0} \\
  
  \midrule
  \rowcolor[HTML]{DADADA}
  NanoHTNet (Ours)      &42.5    &47.7    &44.3    &47.4    &49.3    &55.9    &46.0    &44.2    &55.0    &61.8    &47.8    &45.0    &50.6    &38.1    &39.2    &47.6 \\
  \rowcolor[HTML]{DADADA}
  {NanoHTNet + \blue{PoseCLR}} (Ours) &\textbf{42.2} &{47.4} &\textbf{{43.6}} &\textbf{{47.1}} &\textbf{48.7} &\textbf{55.1} &{45.9} &\textbf{43.5} &\textbf{55.0} &{61.0} &\textbf{{46.9}} &\textbf{45.0} &\textbf{{50.0}} &37.5 &\textbf{38.4} &\textbf{47.1} \\

  \toprule
  \textbf{Protocol \#2 (CPN)} & Dir. & Disc & Eat & Greet & Phone & Photo & Pose & Purch. & Sit & SitD. & Smoke & Wait & WalkD. & Walk & WalkT. & Avg. \\
  \midrule

  Fang \textit{et al.}~\cite{fang2018learning}&38.2 &41.7 &43.7 &44.9 &48.5 &55.3 &40.2 &38.2 &54.5 &64.4 &47.2 &44.3 &47.3 &36.7 &41.7 &45.7 \\

  LCN ~\cite{ci2019optimizing}  &36.9& 41.6& 38.0& 41.0& 41.9 &51.1 &38.2& 37.6& 49.1 &62.1 &43.1& 39.9& 43.5& 32.2& 37.0& 42.2 \\

  Liu \textit{et al.}~\cite{liu2020comprehensive} &35.9 &40.0 &38.0 &41.5 &42.5 &51.4 &37.8 &36.0 &48.6 &56.6 &41.8 &38.3 &42.7 &31.7 &36.2 &41.2 \\

  ST-GCN ~\cite{stgcn}   &36.8 &38.7 &38.2 &41.7 &40.7 &46.8 &37.9 &35.6 &47.6 &51.7 &41.3 &36.8 &42.7 &31.0 &34.7 &40.2 \\

  SRNet~\cite{hrnet}  &35.8 &39.2 &{36.6} &\textbf{36.9} &39.8 &45.1 &38.4 &36.9 &47.7 &54.4 &\textbf{38.6} &36.3 &\textbf{39.4} &30.3 &35.4 &39.4 \\

  MLP-JCG \cite{tang2023mlp} & \textbf{33.7} & {37.4} & 37.3 & {39.6} & 39.8 &47.1 &\textbf{33.7}&\textbf{33.8} &{45.7} &60.5& {39.7} & 37.7& {40.1} &30.1 & {33.8} & 39.3 \\

  MGCN~\cite{mgcn}   &35.7 &38.6 &{36.3} &40.5 &39.2 &44.5 &37.0 &35.4 &46.4 &51.2 &40.5 &35.6 &41.7 &{30.7} &33.9 &39.1 \\
  
  RS-Net \cite{hassan2023regular}  &35.5 &38.3 &\textbf{36.1} &40.5 &39.2 &44.8 &37.1 &34.9 &45.0 &49.1 &40.2 &35.4 &41.5 &31.0 &34.3 &38.9  \\

  \added{GraphMLP~\cite{graphmlp}} &\added{35.1} &\added{38.2} &\added{36.5} &\added{39.8} &\added{39.8} &\added{43.5} &\added{35.7} &\added{34.0} &\added{45.6} &\added{\textbf{47.6}} &\added{39.8} &\added{35.1} &\added{41.1} &\added{\textbf{30.0}} &\added{\textbf{33.4}} &\added{38.4} \\
  
  \midrule
  \rowcolor[HTML]{DADADA}
  NanoHTNet (Ours)   &35.0      &38.4    &36.2    &39.7    &39.3      &43.7      &36.3      &34.5      &44.9    &51.3    &39.6      &35.0    &41.3    &30.9      &33.4      &38.6    \\
  \rowcolor[HTML]{DADADA}
  NanoHTNet + \blue{PoseCLR} (Ours) &{34.7} &\textbf{37.2} &{36.2} &{38.9} &\textbf{{38.4}} &\textbf{42.5} &{36.1} &{34.2} &\textbf{43.9} &{{50.0}} &{39.3} &\textbf{34.8} &{40.4} &{30.8} &\textbf{33.4} &\textbf{38.0} \\

  \toprule
  \textbf{Protocol \#1 (GT)} & Dir. & Disc & Eat & Greet & Phone & Photo & Pose & Purch. & Sit & SitD. & Smoke & Wait & WalkD. & Walk & WalkT. & Avg.\\
  \midrule


 SemGCN~\cite{SemGCN}  &37.8 &49.4 &37.6 &40.9 &45.1 &41.4 &40.1& 48.3& 50.1 &42.2& 53.5 &44.3 &40.5 &47.3& 39.0 &43.8 \\

  ST-GCN ~\cite{stgcn} &{33.4} &39.0 &33.8 &37.0 &38.1 &47.3 &39.5 &37.3 &43.2 &46.2 &37.7 &38.0 &38.6 &30.4 &32.1 &38.1\\

  Liu \textit{et al.}~\cite{liu2020comprehensive}  &36.8 &40.3& 33.0 &36.3 &37.5 &45.0 &39.7 &34.9 &40.3 &47.7& 37.4& 38.5& 38.6& 29.6 &32.0 &37.8 \\

  GraFormer~\cite{GraFormer}  &35.2 &38.0 &30.4 &34.4 &34.7 &43.3 &35.2 &31.4 &38.0 &46.2 &34.2 &35.7 &36.1 &27.4 &30.6 &35.2 \\

  MLP-JCG \cite{tang2023mlp} & \textbf{29.1} & {36.0} & 30.4 & 33.8 & 35.5 & 46.5 & 35.3 & {31.2} & 39.2 & 48.8 & {33.9} & {35.2} & {35.8} & {26.9} & {29.4} & 35.1 \\

  PHGANet \cite{zhang2023learning} &32.4 &36.5 &30.1 &33.3 &36.3 &43.5 &36.1 &30.5 &37.5 &45.3 &33.8 &35.1 &35.3 &27.5 &30.2 &34.9 \\

\added{PGFormer~\cite{pgformer}} &\added{32.3} &\added{36.2} &\added{30.5} &\added{32.6} &\added{33.8} &\added{42.4} &\added{36.3} &\added{29.1} &\added{36.0} &\added{43.9} &\added{33.9} &\added{34.5} &\added{35.1} &\added{26.6} &\added{29.6} &\added{34.2} \\

  \added{GraphMLP~\cite{graphmlp}} &\added{32.2} &\added{38.2} &\added{29.3} &\added{33.4} &\added{33.5} &\added{38.1} &\added{38.2} &\added{31.7} &\added{37.3} &\added{38.5} &\added{34.2} &\added{36.1} &\added{35.5} &\added{28.0} &\added{29.3} &\added{34.2} \\

  HopFIR~\cite{hopfir}  &{31.3} &\textbf{34.0} &28.0 &32.0 &33.1 &42.1 &34.1 &\textbf{28.1} &33.6 &39.8 &31.7 &\textbf{32.9} &33.8 &\textbf{26.7} &\textbf{28.9} &32.7 \\

  Azizi \textit{et al.}~\cite{2025_1} & 30.1& 35.3& 30.6& \textbf{27.6}& 36.2& \textbf{38.4}& \textbf{30.7}& 30.3& 35.9& 40.7& 32.9& 34.9& 35.2& 27.2& 32.0& 32.7 \\
  
  \midrule
  \rowcolor[HTML]{DADADA}
  NanoHTNet(Ours)  &33.5  &36.0  &27.8  &32.4  &31.8   &40.1 &34.6 &28.8  &33.9  &35.1  &31.7  &34.2  &32.8  &27.6  &29.1  &31.9  \\
  
  \rowcolor[HTML]{DADADA}
  NanoHTNet + \blue{PoseCLR} (Ours) &33.3 &35.3 &\textbf{27.6} &{31.9} &\textbf{31.7}  &{{39.6}}  &{33.1} &28.7 &\textbf{33.3} &\textbf{35.0} &\textbf{31.4} &34.0 &\textbf{32.5} &27.6 &\textbf{28.9} &\textbf{31.6} \\

  \toprule
  \end{tabular}
  }
  \label{table:image}
\end{table*}

\section{Experiments}
\VspaceSen
\subsection{Datasets and Evaluation}
\textbf{Human3.6M}~\cite{ionescu2013human3} is the most widely used indoor dataset in the 3D HPE. There are 3.6 million images, 11 professional actors, and 15 different actions in it. Following~\cite{mgcn,zeng2020srnet,GraFormer}, 5 subjects ($S1,S5,S6,S7,S8$) are used for training and 2 subjects ($S9,S11$) are used for testing. For the metrics, we adopt the mean per joint position error (MPJPE) and Procrustes MPJPE (P-MPJPE). The MPJPE (\textit{i.e.}, Protocol 1) measures the Euclidean distance between the ground truth and estimated joints; the P-MPJPE (\textit{i.e.}, Protocol 2) is the MPJPE after aligning the predicted results to the ground truth by translation, rotation, and scale.

\VspaceSen
\VspaceSen
\textbf{MPI-INF-3DHP}~\cite{mehta2017monocular} is also a challenging dataset consisting of 1.3 million images from indoor and outdoor scenes. There are three different scenes in its test set: studio with a green screen (GS), studio without green screen (noGS), and outdoor scene (Outdoor). \added{Following~\cite{mgcn,GraFormer,liu2020comprehensive}, we report metrics of MPJPE, Percentage of Correct Keypoint (PCK), and Area Under Curve (AUC).}

\subsection{Implementation Details}
\textbf{Pre-traing stage.} PoseCLR is pre-trained on the Human3.6M dataset using one NVIDIA RTX 4090 GPU. The 50Hz camera frequency of this dataset means that chronologically adjacent frames are highly correlated. 
\added{If sampled sequentially, these frames would populate the FIFO memory bank with overly similar false negative samples, degrading the contrastive task. 
To mitigate this sampling bias, PoseCLR samples frames at equal intervals of $slice(s)$ to ensure the diversity and quality of negative samples in the memory bank. }
We employ a learning rate of 0.001 and train 200 epochs for effective extraction of human topological prior."

\VspaceSen
\textbf{Fintuning stage.}
With pre-trained parameters, NanoHTNet converges at the optimal performance in 2 epochs, compared to 20 epochs when trained from scratch.
The proposed NanoHTNet consists of 3 stacked spatial and temporal hierarchical mixers.
Notably,  the channel dimension must be a multiple of 48 (\textit{e.g.}, 144, 240, 384) because there are eight heads in the MSA, three split channels in the spatial Hierarchical Mixers, and two split channels in the temporal Hierarchical Mixers.
\added{For a fair comparison, horizontal flip augmentation is adopted referring to ~\cite{poseformer,mixste,strided}. Refinement module is only used in imaged-based comparisons to explore the best performance following our previous work ~\cite{htnet}.}

\VspaceSen
\textbf{Edge Device Deployment.}
To accommodate deployment on edge devices, we evaluate our NanoHTNet on the Jetson Nano. 
Jetson Nano is equipped with a quad-core ARM Cortex-A57 CPU, 4GB memory, and NVIDIA Maxwell with 128 CUDA cores and 0.5 TFLOPs (FP16), offering a small and compact hardware setup ideal for integration into robots.
However, this also imposes specific requirements on the parameters and computational complexity of the algorithms deployed on it.

\begin{table}[t]  
  \centering
  \normalsize
  \renewcommand{\arraystretch}{1.4} 
  \caption{Quantitative comparison of SOTA methods with PoseCLR on Human3.6M under MPJPE and P-MPJPE.}
  \resizebox{\columnwidth}{!}{  
    \begin{tabular}{cl|ll}
    \toprule

    \toprule [1pt]
    &{Method}  &MPJPE ($\downarrow$) &P-MPJPE($\downarrow$)\\
    \midrule
    \multirow{4}[2]{*}{\rotatebox[origin=c]{90}{Image input}} 
    & SemGCN~\cite{SemGCN}   & 60.8 &- \\
    & \cellcolor[HTML]{DADADA}SemGCN + \blue{PoseCLR} & \cellcolor[HTML]{DADADA}58.7  (\blue{$\downarrow$1.1}) &- \cellcolor[HTML]{DADADA}\\
    & MGCN~\cite{mgcn}  & 49.4 &39.1 \\
    & \cellcolor[HTML]{DADADA}MGCN + \blue{PoseCLR}  & \cellcolor[HTML]{DADADA}48.5  (\blue{$\downarrow$0.9}) &\cellcolor[HTML]{DADADA}38.4  (\blue{$\downarrow$0.7})\\
    & GraFormer~\cite{GraFormer}  & 51.8 &- \\
    & \cellcolor[HTML]{DADADA}GraFormer + \blue{PoseCLR} & \cellcolor[HTML]{DADADA}49.9  (\blue{$\downarrow$1.9}) &- \cellcolor[HTML]{DADADA}\\

    \midrule
    \multirow{8}[2]{*}{\rotatebox[origin=c]{90}{Video input}} 
    & MixSTE(T=3)~\cite{mixste}  & 49.6 &38.9 \\
    & \cellcolor[HTML]{DADADA}MixSTE + \blue{PoseCLR} & \cellcolor[HTML]{DADADA}48.7  (\blue{$\downarrow$0.9})& \cellcolor[HTML]{DADADA}38.3 (\blue{$\downarrow$0.6})\\

    & PoseFormerV2 (T=27, RF=243)~\cite{PoseFormerV2}  & 45.2 &35.6 \\
    & \cellcolor[HTML]{DADADA}PoseFormerV2 + \blue{PoseCLR} & \cellcolor[HTML]{DADADA}44.8  (\blue{$\downarrow$0.4}) & \cellcolor[HTML]{DADADA}35.1  (\blue{$\downarrow$0.5})\\
    
    & STCFormer (T=27)~\cite{STCFormer}  & 44.1 &34.7 \\
    & \cellcolor[HTML]{DADADA}STCFormer + \blue{PoseCLR} & \cellcolor[HTML]{DADADA}43.5  (\blue{$\downarrow$0.6})& 34.3  (\blue{$\downarrow$0.4}) \cellcolor[HTML]{DADADA}\\

    & NanoHTNet (Ours, T=27, RF=243)  & 43.5 &34.5 \\
    & \cellcolor[HTML]{DADADA}NanoHTNet + \blue{PoseCLR} & \cellcolor[HTML]{DADADA}43.0  (\blue{$\downarrow$0.5}) & 34.1  (\blue{$\downarrow$0.4})\cellcolor[HTML]{DADADA}\\
    \bottomrule
    \end{tabular}%
  }
  \label{tab:poseclr}%
\end{table}%

\subsection{Comparison with State-of-the-art Results}
\textbf{Compared with Video-based Methods.}
As shown in Tab.~\ref{table:video}, existing Seq2Frame and Seq2Seq approaches fail to reach real-time speed (30 FPS) on Jetson Nano.
For the same video sequence input,
Seq2Frame methods output only intermediate frames, achieving better performance but slower inference. 
Seq2Seq methods match the input length, offering higher efficiency but lower performance. 
To be competitive, Seq2Seq methods often require extensive computation (MixSTE, and STCFormer reach 3420 and 4346 MFLOPs, respectively).
In this case, our NanoHTNet based on Seq2Seq pipeline leverages spatio-temporal priors to enhance structural understanding and significantly improves efficiency in processing skeleton data.
Despite having only 1.52MB in parameters and 33 MFLOPS, NanaHTNet achieves impressive accuracies of 44.5\textit{mm} MPJPE and 35.2\textit{mm} P-MPJPE.
By extending input frames (9$\to$27) and channel dimension (240$\to$384), NanoHTNet-L outperforms other State Of The Art (SOTA) methods in the speed-accuracy trade-off, achieving 43.5\textit{mm} MPJPE, 34.5\textit{mm} P-MPJPE with only 4.4MB parameters and 156 MFLOPs.
Moreover, its inference speed on Jetson Nano further reach 112.76 FPS, satisfying real-time inference needs.

\begin{table}[t]
    \centering
    \renewcommand{\arraystretch}{1.3} 
    \caption{
	    Quantitative comparison on MPI-INF-3DHP with five metrics. Here, $\uparrow$ indicates the higher, the better. 
	}
    \begin{tabular}{l|cccc|c}
        \toprule 
        \toprule[1pt]
        \multirow{2}{*}{Methods} & \multicolumn{4}{c|}{\added{PCK}} & \multirow{2}{*}{AUC} \\
        
        \cmidrule(lr){2-5}
        & GS & no GS & Outdoor & All & \\
        \midrule

        Martinez \textit{et al.}~\cite{simplebaseline}   & 49.8 &42.5 &31.2 &42.5 & 17.0 \\
        LCN ~\cite{ci2019optimizing}   & 74.8 &70.8 &77.3 &74.0 &36.7 \\
        Zhou \textit{et al.}~\cite{zhou2017towards}  & 75.6 &71.3 &80.3 &75.3 &38.0 \\
        SRNet~\cite{zeng2020srnet}    &- &-  &80.3 &77.6 &43.8  \\
        Liu \textit{et al.}~\cite{liu2020comprehensive}   &77.6 &80.5 &80.1 &79.3 &45.8 \\
        Zeng \textit{et al.}~\cite{zeng2021learning}   &- &-  &84.6 &82.1 &46.2  \\
        Xu \textit{et al.}~\cite{xu2021graph}  &81.5 &81.7 &75.2 &80.1 &45.8 \\

    \added{PGFormer~\cite{pgformer}} &\added{84.4} &\added{84.5} &\added{77.4} &\added{83.9} &\added{52.3}\\

        UGRN \cite{li2023pose} & 86.2 & 84.7 & 81.9 & 84.1 & 53.7 \\

        RS-Net \cite{hassan2023regular} &- &- &- &85.6 &53.2  \\
        
        MGCN~\cite{mgcn}  &{86.4} &{86.0} &{85.7} &{86.1} &{53.7} \\
        \rowcolor[HTML]{DADADA}
        MGCN + \blue{PoseCLR}  &{87.1} &{86.6} &{86.7} &{86.4} &{54.4} \\
    \added{GraphMLP~\cite{graphmlp}} &\added{87.3} &\added{87.1} &\added{86.3} &\added{87.0} &\added{54.3}\\
        HopFIR~\cite{hopfir}   &{89.1} &{85.9} &{85.9} &{87.2} &\textbf{57.0} \\
    \midrule
    \rowcolor[HTML]{DADADA}
	NanoHTNet(Ours) &88.9 &86.2 &85.9 &86.7  &56.1	\\
    \rowcolor[HTML]{DADADA}
 	NanoHTNet + \blue{PoseCLR} &\textbf{89.4} &\textbf{86.6} &\textbf{86.1} &\textbf{88.1}  &\textbf{57.0}	\\
	\bottomrule
	\end{tabular}
	\label{tab:3dhp}
\end{table}

\VspaceSen
\textbf{Compared with Image-base methods.}
We compare our NanoHTNet with state-of-the-art image-based methods on Human3.6M.
As shown in Tab.~\ref{table:image} (top part), the 2D keypoints from the 2D detector CPN~\cite{chen2018cascaded} are used as input, and our NanoHTNet achieves the best results under both MPJPE (47.6\textit{mm}) and P-MPJPE (38.6\textit{mm}). 
Moreover, MPJPE can be further improved to 47.1\textit{mm} under MPJPE and 38.0\textit{mm} under P-MPJPE after introducing PoseCLR.
In particular, we can observe that NanoHTNet with PoseCLR obtains the best performance in all kinds of action in MPJPE. 
As shown in Tab.~\ref{table:image} (bottom part), 
NanoHTNet with PoseCLR also reaches the best performance under MPJPE (31.9\textit{mm}) using ground truth 2D joints, outperforming~\cite{hopfir} by 1.1\textit{mm}.

\VspaceSen
\textbf{Methods with PoseCLR.}
To validate the generality and effectiveness of PoseCLR as a pre-training approach in 3D HPE task, we employ three image-based methods (SemGCN~\cite{SemGCN}, MGCN~\cite{mgcn}, GraFormer~\cite{GraFormer}), three video-based methods (MixSTE~\cite{mixste}, PoseFormerV2~\cite{PoseFormerV2}, STCFormer~\cite{STCFormer}), and our NanoHTNet as the encoder for training and testing. 
As indicated in Table~\ref{tab:poseclr}, the introduction of PoseCLR leads to further improvements in the MPJPE metric across all methods (\textit{e.g.}, 3.67\% increase in the MPJPE metric for GraFormer after incorporating PoseCLR). 
Furthermore, the training time of finetuning is approximately one-tenth of that required for training from scratch. 
This indicates that pre-training with PoseCLR enables better initialization parameters embedded with inherent priors of human body topology, which significantly enhance the accuracy and robustness of understanding the human body for 3D HPE models.

\VspaceSen
\textbf{Comparison on MPI-INF-3DHP.}
To evaluate the robustness of our NanoHTNet and PoseCLR, we only train the model on the Human3.6M and test it on more challenging MPI-INF-3DHP directly, without any extra training and refinement.
As shown in Tab.~\ref{tab:3dhp}, our NanoHTNet achieves the best performance on all various scenes. Additionally, loading pre-trained parameters from Human3.6M into models (MGCN and NanoHTNet) facilitates further performance enhancements.
This verifies the strong generalization of proposed NanoHTNet and PoseCLR in unseen environments.

\begin{table}[t]
    \footnotesize
    \centering
    \setlength{\tabcolsep}{1.80mm} 
    \renewcommand{\arraystretch}{1.3} 
    \caption{Ablation study for components and structures. The evaluation is performed on Human3.6M. We set NanoHTNet with only GBI sub-module and ETST as Baseline.}
	\begin{tabular*}{\columnwidth}{l|cccc|ccc}
		\toprule
		& LJC & \makecell IPC & \makecell{GBI}  & ETST & \makecell{Params \\ (M)} & \makecell{\added{FLOPS} \\ \added{(M)}} & MPJPE \\
		\midrule
		Baseline &   &   &\added{\Checkmark}         &  	& \added{28.62} &  \added{762.2} & \added{47.4}\\
             &   &   &\Checkmark         &\Checkmark  	& 1.25  & \added{52.1} & 47.1\\
		&\Checkmark  &   &               &\Checkmark    & 1.47  & \added{39.7} & 46.2\\
		& \Checkmark & & \Checkmark          &\Checkmark    & 1.43  & \added{45.9} & 45.9\\
		&        & \Checkmark  & \Checkmark      &\Checkmark    & 1.74  & \added{39.6} & 46.5\\
		& \Checkmark       & \Checkmark   &    &\Checkmark    & 1.92  & \added{30.1} & 45.0\\
            & \Checkmark       & \Checkmark   &     \Checkmark   &    & 31.75  & \added{501.2} & 44.1\\
            \midrule
            Parallel &\Checkmark    &\Checkmark  & \Checkmark  & \Checkmark  &1.52  & \added{33.4} &46.0\\
            \added{Parallel} &\added{\Checkmark}    &\added{\Checkmark}  & \added{\Checkmark}  &  &\added{31.75}  & \added{501.2} &\added{45.2}\\
		Serial &\Checkmark    &\Checkmark  & \Checkmark   & \Checkmark  &4.07  & \added{107.7} &44.7\\
            \added{Serial} &\added{\Checkmark}    &\added{\Checkmark}  & \added{\Checkmark}   &  &\added{39.58}  & \added{612.9} &\added{44.3}\\
		\midrule
		\rowcolor[HTML]{DADADA}  
		NanoHTNet    & \Checkmark & \Checkmark & \Checkmark & \Checkmark  & 1.52  & \added{33.4} & 44.5\\
		\bottomrule  
	\end{tabular*}
	\label{tab:ablation_component}
\end{table}

\begin{table}[t]
	\centering
	\small
        \renewcommand{\arraystretch}{1.3}
        \caption{Ablation study on RF (receptive field, input frames before DCT and LF) and Frames (input frames after DCT and LF). The evaluation is implemented on Human3.6M.}

		\begin{tabular}{cc|ccc}
			\toprule
			Frames ($t$) & RF ($n$)   &Params (M)  & FLOPs (M) &MPJPE\\
			\midrule
			1      & 9               & 1.39   &19.6 &47.3  \\
			3      & 27              & 1.40   &22.4  &46.8  \\
			9      & 27              & 1.42   &29.9  &46.4  \\
			9      & 81              & 1.44   &30.7 &45.2     \\
                27      & 81              & 1.48   &53.2 &45.0     \\
                \rowcolor[HTML]{DADADA}
			9      & 243             & 1.52   &33.4  &44.5  \\
			27     & 243 		     & 1.56   &55.8  &44.5 \\
			81     & 243 		     & 1.66   &123.2  &45.1 \\

			\bottomrule
		\end{tabular}
	\label{tab:ablation_rf}
\end{table}

\begin{figure}[t]
    \centering
    \includegraphics[width=1.0\linewidth]{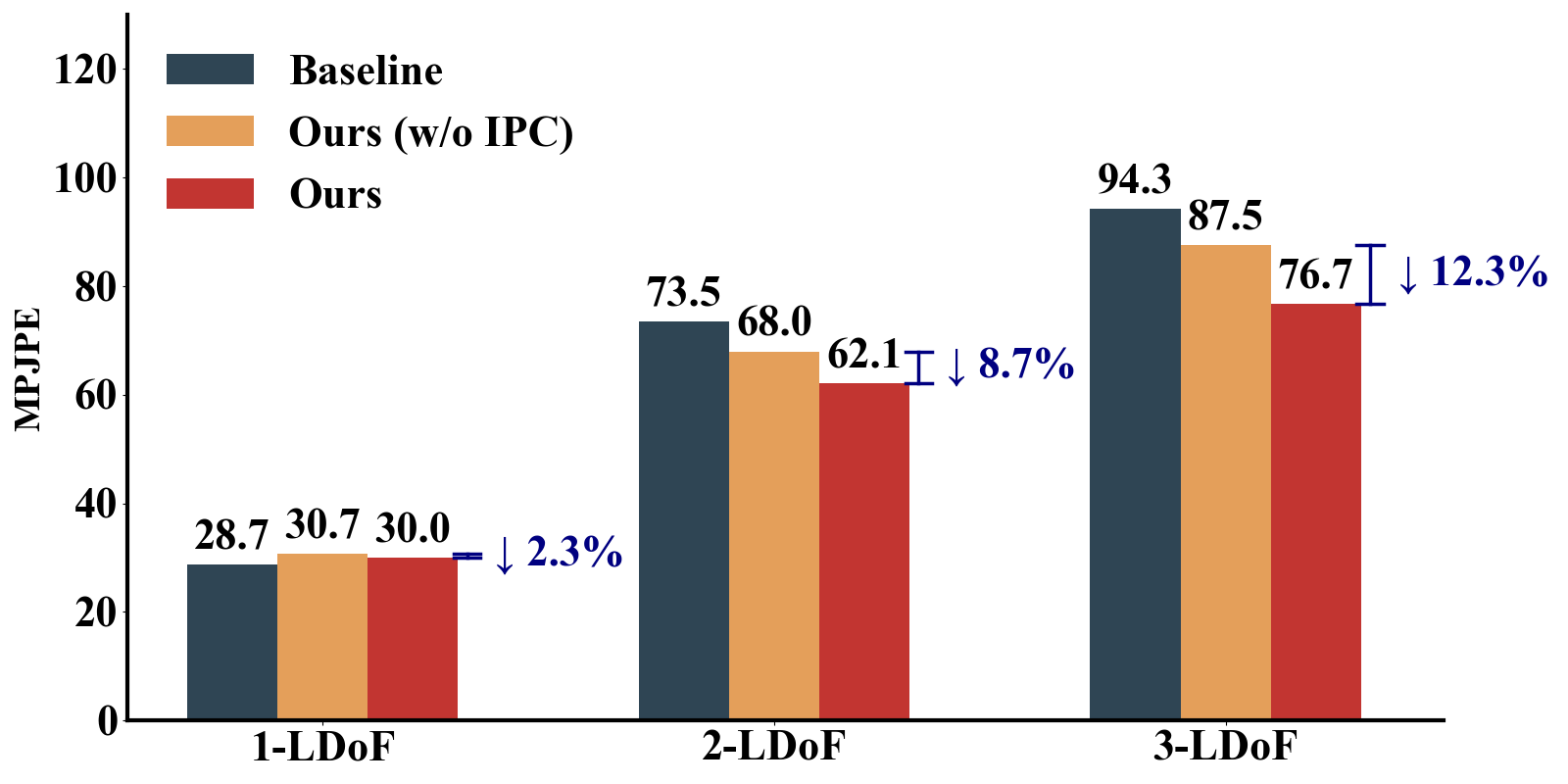}
    \caption
    {
        \added{Ablation study for the effect of IPC module.}
    }
    \label{fig:ablation}
   \VspacePa
    
\end{figure}

\begin{table}[t]
	\centering
	\small
	\renewcommand{\arraystretch}{1.3}
	\caption{Ablation study on hyperparameters of PoseCLR. Views ($\alpha$) is the number of cameras. Memory ($\beta$) is the size of the memory bank. Slice ($s$) is the sampling interval. Decay ($\tau$) is the rate of momentum update.}
	\begin{tabular}{cccc|c}
		\toprule
            \footnotesize Views($\alpha$) & \footnotesize Memory Bank($\beta$) & \footnotesize Slice($s$) & \footnotesize Decay($\tau$) & \footnotesize MPJPE ($\downarrow$)\\
		\midrule
            \rowcolor[HTML]{DADADA}
		4       & 32768            & 3          & 0.999      & 43.7 \\
		3       & 32768            & 3          & 0.999      & 44.1 \\
		2       & 32768            & 3          & 0.999      & 44.5 \\
		\midrule
		4       & 65536            & 3          & 0.999      & 43.7 \\
		4       & 16384            & 3          & 0.999      & 43.8 \\
		4       & 8192             & 3          & 0.999      & 44.0 \\
		\midrule
		4       & 32768            & 1          & 0.999      & 45.1 \\
		4       & 32768            & 2          & 0.999      & 43.9 \\
		4       & 32768            & 4          & 0.999      & 44.2 \\
		\midrule
		\added{4}       & \added{32768}            & \added{3}          & \added{1.0}        & \added{45.3} \\
		\added{4}       & \added{32768}            & \added{3}          & \added{0.99}       & \added{43.8} \\
		\added{4}       & \added{32768}            & \added{3}          & \added{0.9}        & \added{44.3} \\
		\bottomrule
	\end{tabular}
	\label{tab:poseclr_ablation}
\end{table}

\begin{figure*}[t]
    \centering
    \includegraphics[width=1.0\linewidth]{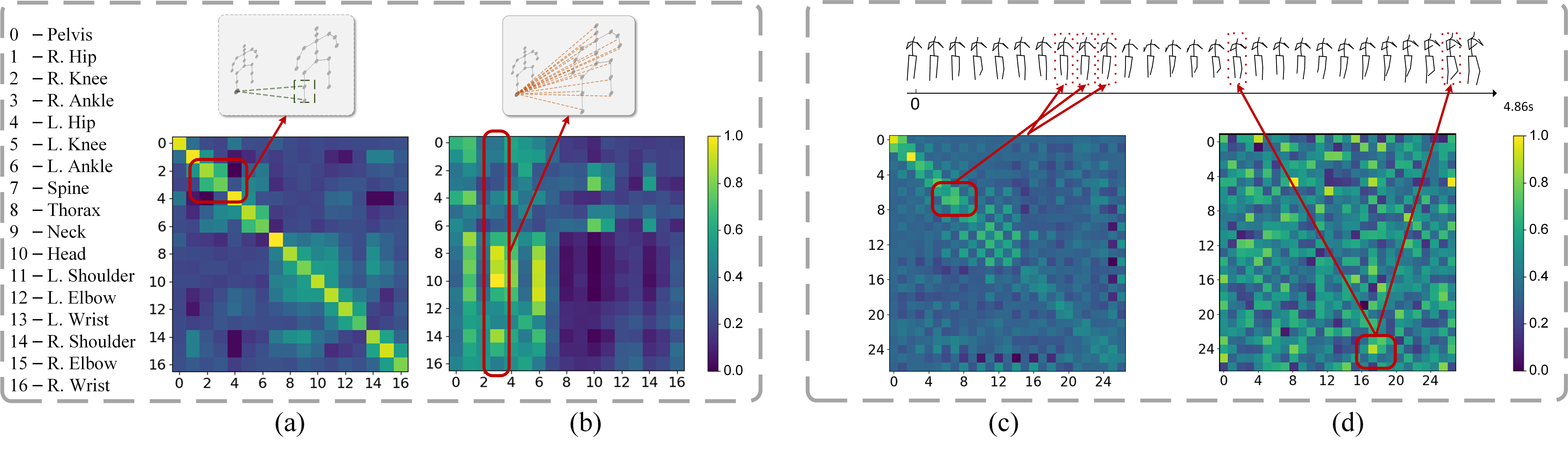}
    \VspaceFa
    \VspaceFa
    \caption
    {
        (a) (b) Visualization of adjacency matrix from LJC and self-attention maps from GBI. The x-axis and y-axis correspond to the query of 17 joints. 
        (c) (d) Visualization of adjacency matrix from IME and self-attention maps from GCP. The x-axis and y-axis correspond to the query of 27 frames. 
        Yellow indicates stronger attention. 
        Attention values are normalized from 0 to 1. 
    }
    \label{fig:viz}
\end{figure*}

\begin{figure*}[t]
    \centering
    \includegraphics[width=0.95\linewidth]{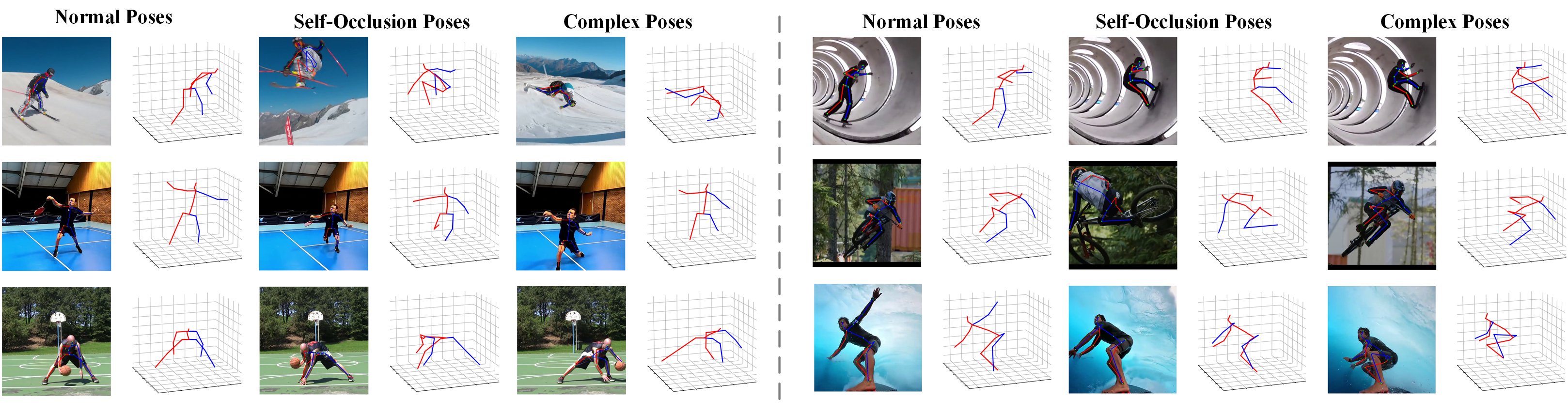}
    \VspaceFb
    \caption
    {
        Qualitative results of NanoHTNet in wild scenes. 
    }
    \label{fig:wild}
    \VspaceFa
\end{figure*}

\VspacePc
\subsection{Ablation Study}
\textbf{Impact of Model Components.}
Given the similar structure of the Spatial and Temporal Hierarchical Mixers, Spatial Hierarchical Mixer is selected as a representative to investigate the impact of model components.
As shown in the top part of Tab.~\ref{tab:ablation_component}, NanoHTNet with only one single-level module cannot achieve satisfactory performance, while the combination of all proposed modules performs the best via learning representations from various levels. 
Notably, IPC cannot work individually due to no operations on 0-LDoF and 1-LDoF joints. To further investigate the influence of IPC, we categorize the limb joints by LDoFs and calculate the average MPJPE of joints in each category in Fig.~\ref{fig:ablation}.
It is obvious to find that the primary reduction in estimation errors originates from distal limb joints (MPJPE on 1-LDoF, 2-LDoF and 3-LDoF joints noticeably reduces by \added{ $2.3\%$, $8.7\%$ and $12.3\%$}, respectively), which can be attributed to the increased number of topological constraints that these joints receive as detailed in Sec.~\ref{sec:Mixers}:
\textit{(i)} 3-LDoF joints are constrained by both 1-LDoF and 2-LDoF joints; 
\textit{(ii)} 2-LDoF joints are only constrained by 1-LDoF joints; \textit{(iii)} 1-LDoF joints are not subject to any constraints. 
Moreover, we observe that model parameters significantly reduce via the introduction of the ETST (31.75M $\rightarrow$ 1.52M) while maintaining performance, which is highly beneficial for deployment on edge devices.
In summary, the efficiency of NanoHTNet is attributed to two key factors: (1) ETST with efficient data mapping significantly reduces parameters and computational complexity; (2) The hierarchical design could capture multi-level spatio-temporal priors, enabling effective understanding of the human body and superior estimation with smaller model size.

\begin{table}[t]  
  \centering
  \footnotesize
  \renewcommand{\arraystretch}{1.5} 
  \caption{Ablation study on hyperparameters of NanoHTNet. Layers refer to the number of stacked Mixers. Dimensions refer to the input channel. The evaluation is implemented on Human3.6M.}
  \resizebox{\columnwidth}{!}{  
    \begin{tabular}{cc|ccc}
    \toprule
    Layers (\textit{L}) &Dimensions (\textit{C}) &Params (M)  &FLOPs (M)  &MPJPE \\
    \midrule
    \rowcolor[HTML]{DADADA}
     3  &240 &1.52   &33.4   &44.5 \\
     2  &240  &1.07  &24.0 &45.1\\
     4  &240  &1.97  &44.7 &44.8\\
     3  &144  &0.83  &18.4 &45.6\\
     3  &384  &2.76  &60.6 &44.7\\
    \bottomrule
    \end{tabular}%
  }
  \label{tab:ablation_hyper}%
\end{table}%

\begin{figure}[t]
    \centering
    \includegraphics[width=1.0\linewidth]{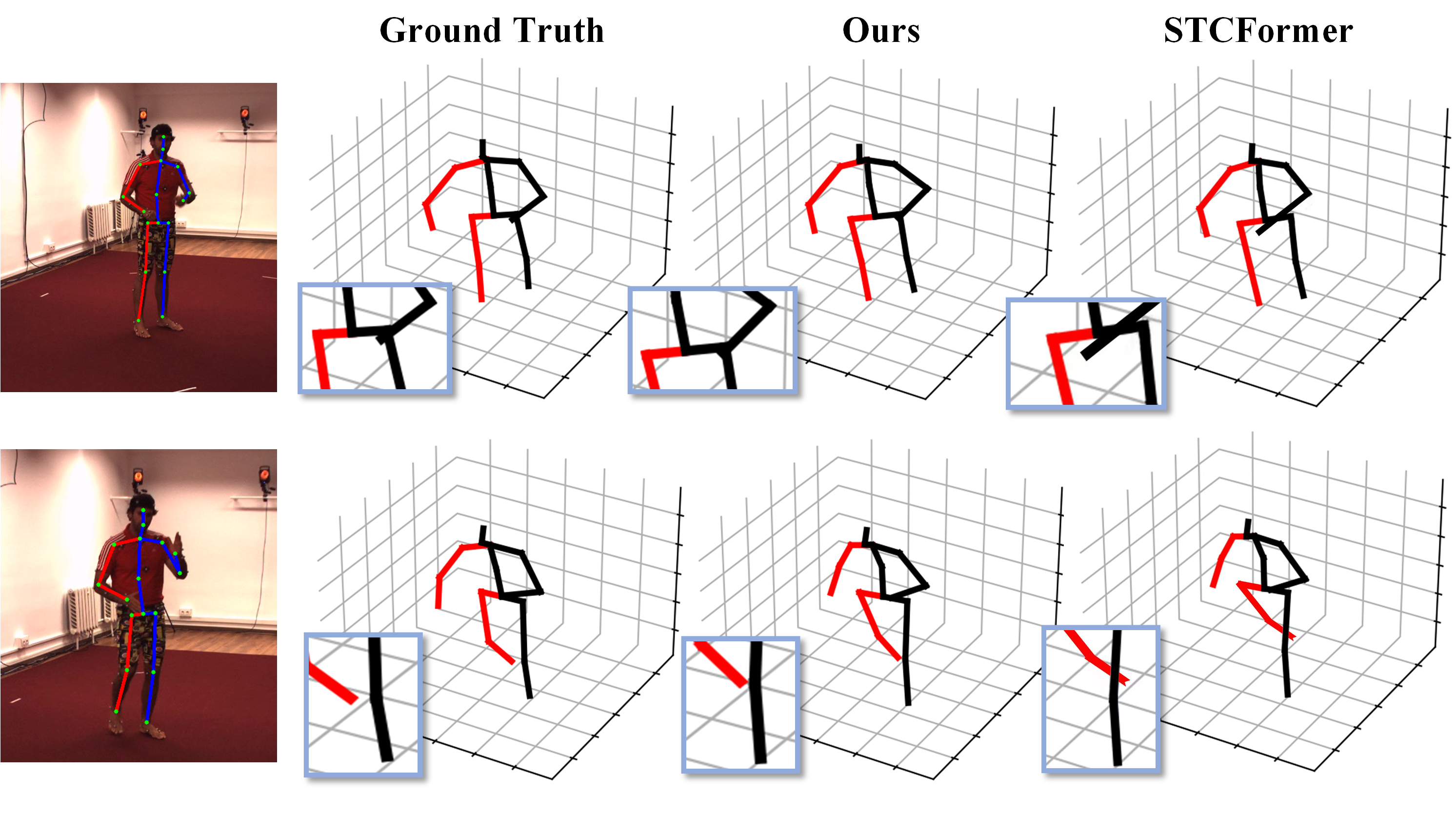}
    \caption
    {
        Qualitative comparison between our method, previous SOTA method, and ground truth on Human3.6M.
    }
    \label{fig:compare}
    \VspacePa
\end{figure}

\VspaceSen
\textbf{Impact of Model Structures.}
As for the structures (middle part of Tab.~\ref{tab:ablation_component}), the serial structure exhibits competitive performance (46.0\textit{mm}) due to the local-to-global learning; the parallel structure can reduce parameters from 4.07M to 1.52M due to the channel-split structure. 
The channel-split progressive structure of NanoHTNet adopts residual connections and combines the advantages of these two designs, which performs the best (44.5\textit{mm}) and maintains a small model size (1.52M).

\VspaceSen
\textbf{Impact of Frames and Receptive Field.}
As illustrated in Fig.~\ref{tab:ablation_rf}, the receptive field significantly affects the overall performance by capturing extensive spatio-temporal context, while input frames determine the parameters and FLOPs.
In this work, the introduction of ETST enables NanoHTNet to maintain a tiny model size, and frequency low-pass filtering prevents size boosting as the receptive field expands.
For example, when the receptive field increases (27$\rightarrow$81) with the same input length (9 frames), NanoHTNet achieves 1.2\textit{mm} improvement under MPJPE while the parameters and FLOPs merely increase 0.02M (1.42M$\rightarrow$1.44M) and 0.8M (29.9M$\rightarrow$30.7M), respectively.
Notably, with larger receptive fields and more input frames, the performance bottleneck shifts towards the model size, prompting further exploration into how these hyperparameters impact performance.

\VspaceSen
\textbf{Impact of Hyperparameters in PoseCLR.}
We evaluate different hyperparameter settings of PoseCLR in ~\ref{tab:poseclr_ablation}. 
We observe that: 
(i) More views ($s=4$) are crucial for pre-training; 
(ii) The increase of memory bank $\beta$ brings limited improvement to the MPJPE metric, thus $\beta$ is set to 32,768 to balance memory usage and performance;
(iii) Sampling interval $s$ effectively mitigates the noise from adjacent frames as negative samples, yet a large interval reduces the training database. PoseCLR performs optimally at $s=3$; 
\added{(iv)  The momentum update decay rate $\tau$ is optimal at 0.999, as high or low values degrade performance by disrupting learning stability.}

\VspaceSen
\textbf{Impact of Hyperparameters in NanoHTNet.}
As shown in Tab.~\ref{tab:ablation_hyper}, we explore the hyperparameter settings of NanoHTNet. 
It can be seen that the channel dimension $C$ and layers $L$ of the hierarchical mixing block have a significant impact on the performance, and the combination of $C = 240$ and $L = 3$ works best. 
The reason for this is the inadequacy of parameters in NanoHTNet for capturing the complex topological features of the human body, leading to underfitting, while an overly large model tends to overfit the training data, focusing on noise instead of achieving generalization.

\subsection{Visualization}
\VspaceSen
\textbf{Attention Visualization.}
In order to illustrate the effectiveness of Hierarchical Mixers, we visualize the adjacency matrix from LJC and IME, along with multi-head self-attention maps from GBI and GCP. 
It can be observed in Fig.\figeight{fig:viz}{a} that the joints (\textit{e.g.}, Joint 3, Right Ankle) primarily pay attention to themselves and their physically connected ones. This demonstrates that LJC in NanoHTNet mainly focuses on the most related local region. 
In addition, Fig.\figeight{fig:viz}{b} shows that the limb joints (\textit{e.g.}, Joint 3, Right Ankle) concentrate on the other limb joints across limbs, which demonstrates the GBI mainly learns the less adjacent global context among joints. 
Similarly, Fig.\figeight{fig:viz}{c} demonstrates that in the IME module, each frame focuses on the local instantaneous features of adjacent frames (\textit{e.g.}, the attention in frame 7 is primarily on frames 8 and 9). 
Fig.\figeight{fig:viz}{d} illustrates that each frame pays more attention to the global coherence and coordination of actions in the GCP module (\textit{e.g.}, the attention in frame 15 is on frame 25 due to the cyclical repetition of similar actions).
Overall, these results prove that our Hierarchical Mixers can effectively extract multiple semantic level features according to the spatio-temporal hierarchy of the human topology. 
Additionally, different sub-modules sequentially extract features at specific target levels, which aligns well with our original design intent.

\VspaceSen
\textbf{Visualization in wild scenes.}
As illustrated in Fig.~\ref{fig:wild}, the following indoor and outdoor sports are selected from online video websites for evaluation: alpine skiing, badminton, basketball, skateboarding, mountain biking, and surfing.
Although some complex and self-occlusion poses in these images are unseen or rare on Human3.6M, our NanoHTNet can produce accurate and plausible 3D poses owning to the comprehensive understanding of human spatial-temporal topology.

\VspaceSen
\textbf{Qualitative comparison of Visualization.}
Fig.~\ref{fig:compare} provides qualitative comparisons on Human3.6M between NanoHTNet and the state-of-the-art approach,  \textit{i.e.}, \added{STCFormer~\cite{STCFormer}}.
It can be seen that our method produces more precise 3D keypoints at the ends of limbs, demonstrating its effectiveness.

\section{Conclusion}
In this work, we leverage spatio-temporal priors of human body to improve model efficiency in 3D human pose estimation.
First, we propose a nano network namely NanoHTNet based on the explicit human features, which captures spatial hierarchical features and temporal multi-scale kinematics of the human topology. 
Furthermore, we integrate Efficient Temporal-Spatial Tokenization and Low-pass Filtering to achieve more efficient data processing and computation.
Second, further based on cross-view spatio-temporal topological consistency, a contrastive learning approach PoseCLR is proposed to capture implicit features of the human body.
Extensive experiments validate that NanoHTNet achieves outstanding performance and efficiency, enabling real-time inference on computationally constrained devices like Jetson Nano.
Moreover, as a general pre-training approach for the 3D HPE, PoseCLR effectively optimizes the initialization parameters for various encoders, further improving performance.

\bibliographystyle{IEEEtran}

\begin{IEEEbiography}
[{\includegraphics[width=1in,height=1.25in,clip,keepaspectratio]
{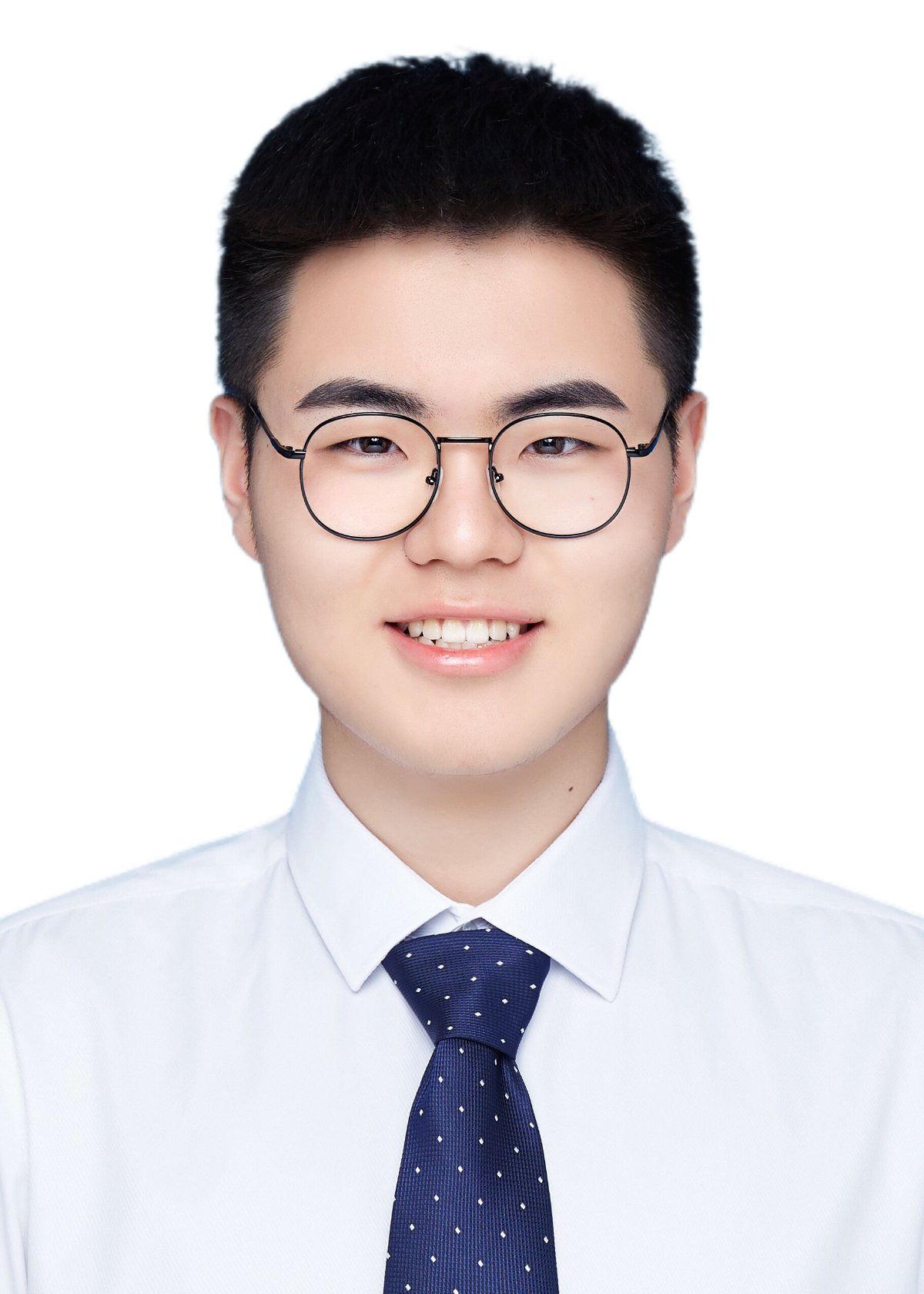}}]{Jialun Cai} received the B.Eng. degree from Central South University in 2021.
He graduated from Computer Application Technology at the School of Electronic and Computer Engineering, Peking University (PKU), China, under the supervision of  Prof. Hong Liu.
He is currently an algorithm engineer at Ant Group.
His research interests lie in 3D human pose estimation, deep learning, and their applications and deployments to computer vision.
\end{IEEEbiography}

\begin{IEEEbiography}[{\includegraphics[width=1in,height=1.25in,clip,keepaspectratio]{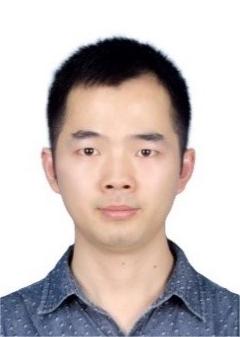}}]{Mengyuan Liu} is an Assistant Professor at Peking University. He received a Ph.D. degree in 2017 under the supervision of Prof. H. Liu from the School of EE\&CS, Peking University (PKU), China. His research interests include human action recognition, human motion prediction, and human motion generation using RGB, depth, and skeleton data. Related methods have been published in TIP, TCSVT, T MM, PR, CVPR, ECCV, ACM MM, and AAAI. He has been invited to be a Technical Program Committee (TPC) member for ACPR, ACM MM, and AAAI.
\end{IEEEbiography}

\begin{IEEEbiography}[{\includegraphics[width=1in,height=1.25in,clip,keepaspectratio]{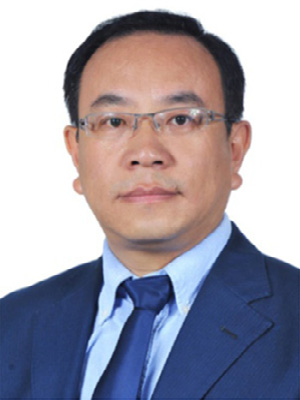}}]{Hong Liu} is a full professor at Peking University, and serves as Co-director of the Center of Embodied Intelligence and Robotics, Institute of AI, Peking University and is also the founding dean of Chongqing Liangjiang School of Artificial Intelligence. He serves as the vice president of the Chinese Association for Artificial Intelligence (CAAI), and the founding Editor-in-Chief for the top international journal of CAAI Transactions on Intelligence Technology.  He earned the National Aerospace Science and Technology Progress Award, Geneva International Invention Award, the First Price in Nature Science of Wu Wenjun Artificial Intelligence Science and Technology Award, the First Price in Natural Science of Shenzhen Science and Technology Award. Hong Liu has engaged in the research of artificial intelligence, robotics, machine Learning and intelligent human robot interaction for more than twenty years. He published over 300 papers in international journals and proceedings in the above areas and these papers have been cited over 14500 times.
\end{IEEEbiography}

\begin{IEEEbiography}[{\includegraphics[width=1in,height=1.25in,clip,keepaspectratio]{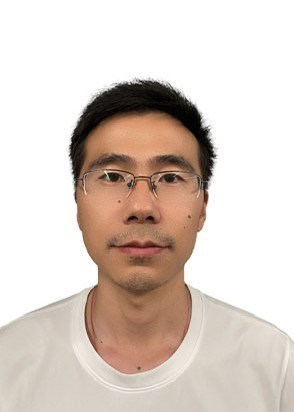}}]{Shuheng Zhou}  received the Master degree from East China University of Science and Technology. He is currently a Senior Algorithm Expert at Ant Group, with extensive experience in the application and deployment of deep learing algorithms. His primary research focus is on the inference acceleration and deployment of Large Language Models (LLMs).
\end{IEEEbiography}

\begin{IEEEbiography}[{\includegraphics[width=1in,height=1.25in,clip,keepaspectratio]{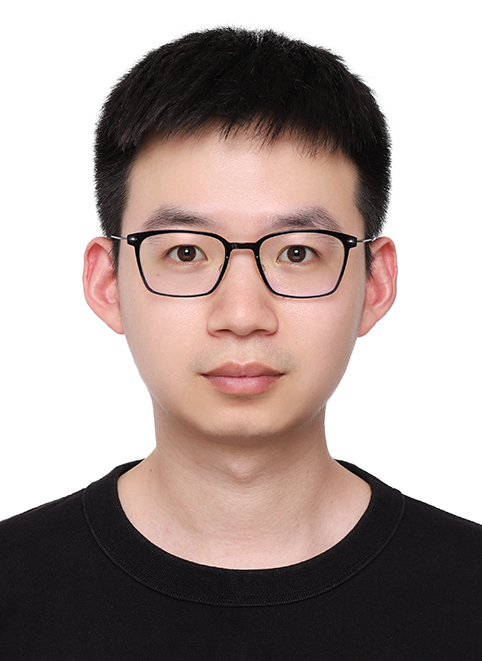}}]{Wenhao Li} is currently a Postdoctoral Researcher at the School of Computer Science and Engineering, Nanyang Technological University, Singapore.  
He received the Ph. D. degree from the School of Computer Science, Peking University, China. 
His research interests lie in deep learning, machine learning, and their applications to computer vision. computer vision. 
\end{IEEEbiography}

\vfill 
\end{document}